\documentclass[journal,twoside,web]{ieeecolor}
\usepackage{generic}
\usepackage{cite}
\usepackage{amsmath,amssymb,amsfonts}
\usepackage{url}
\usepackage{booktabs}
\usepackage{xcolor}
\usepackage{multirow}
\usepackage{makecell}
\usepackage{subcaption}
\usepackage{rotating}
\usepackage{pdflscape}
\usepackage{algorithmic}
\usepackage{graphicx}
\usepackage{algorithm,algorithmic}
\usepackage{hyperref}
\hypersetup{hidelinks=true}
\usepackage{textcomp}
\usepackage[rgb]{xcolor}
\def\BibTeX{{\rm B\kern-.05em{\sc i\kern-.025em b}\kern-.08em
    T\kern-.1667em\lower.7ex\hbox{E}\kern-.125emX}}
\begin{document}
\title{Hierarchical Vision Transformer with Prototypes for Interpretable Medical Image Classification}

\author{Luisa Gallée, Meinrad Beer, Michael Götz
\thanks{
This research was supported by the German Federal Ministry of Research, Technology and Space (BMFTR) within RACOON COMBINE "NUM 2.0" (FKZ: 01KX2121).
}
\thanks{Luisa Gallée is with the Experimental Radiology, Ulm University Medical Center, Ulm, Germany and XAIRAD - Cooperation for Artificial Intelligence in Experimental Radiology, Ulm, Germany (e-mail: luisa.gallee@uni-ulm.de).}
\thanks{Meinrad Beer is with the Department of Diagnostic and Interventional Radiology, Ulm University Medical Center, Ulm, Germany, XAIRAD - Cooperation for Artificial Intelligence in Experimental Radiology, Ulm, Germany, and i2SouI - Innovative Imaging in Surgical Oncology Ulm, Ulm University Medical Center, Ulm, Germany (e-mail: meinrad.beer@uniklinik-ulm.de).}
\thanks{Michael Götz is with the Experimental Radiology, Ulm University Medical Center, Ulm, Germany, XAIRAD - Cooperation for Artificial Intelligence in Experimental Radiology, Ulm, Germany, and i2SouI - Innovative Imaging in Surgical Oncology Ulm, Ulm University Medical Center, Ulm, Germany (e-mail: michael.goetz@uni-ulm.de).}}
\maketitle

\begin{abstract}
Explainability is an important requirement for machine learning models in high-risk domains such as medicine. We introduce HierViT, a hierarchical Vision Transformer that combines competitive predictive performance with multiple mechanisms to support the interpretation of its predictions. The model first predicts clinically relevant visual appearance attributes, which are then used to infer the final target prediction through a hierarchical reasoning process. To facilitate understanding of the intermediate representations, attribute-specific prototypes are visualized as representative image examples, while attribute-specific attention maps highlight image regions associated with each visual feature. Together, these attribute-level components provide complementary views of the model's reasoning by linking representative image examples and spatial attention to semantically defined visual features. We evaluate HierViT on two medical benchmark datasets: LIDC-IDRI for lung nodule assessment and derm7pt for skin lesion classification. Across both datasets, HierViT achieves competitive predictive performance while providing attribute-level prototypes and attention visualizations that support the interpretation of its predictions.
\end{abstract}


\section{Introduction}
\label{sec:introduction}

Since their adaptation from NLP to computer vision in 2021, Vision Transformers (ViTs) have revolutionized image processing, excelling in areas like image segmentation and self-supervised learning \cite{dosovitskiy2021vit,Kirillov_2023_ICCV,NEURIPS2021_64f1f27b,Caron_2021_ICCV,He_2022_CVPR}. 
In the field of explainable AI, ViTs inherently provide an initial insight into the model's logic through self-attention extraction \cite{dosovitskiy2021vit}. However, in high-risk domains like medicine, interpretability must extend beyond visual attention alone \cite{rudin2019stop}. 
Although attention heatmaps localize regions that influence a prediction, they do not provide semantic explanations of the visual features that make these regions informative.

Although a range of explainability methods have been proposed and extended from Convolutional Neural Networks (CNNs) to ViTs, existing approaches typically focus on a single explanation modality, rather than combining multiple complementary mechanisms.
One promising line of explainable AI research is hierarchical modeling, which aligns well with human decision-making by structuring predictions based on step-by-step reasoning \cite{koh2020concept,lalonde_encoding_2020}. By breaking prediction into sequential stages, first identifying expert-defined features and then performing the final classification, these models mirror the stepwise logic of human decision-making and enhance both transparency and interpretability.
Also, prototype-based models gained significant impact in the field of explainable AI, providing tangible, case-based examples to ground the model's logic \cite{NEURIPS2019_adf7ee2d,gallee2023interpretable}. With each of these approaches offering individual advantages, a recent trend is to combine those approaches to benefit from the complementary aspects of each interpretable tool  \cite{gallee2023interpretable,reyes2020interpretability}. 

\begin{figure}[!htbp]
    \centering
    \includegraphics[width=0.8\linewidth]{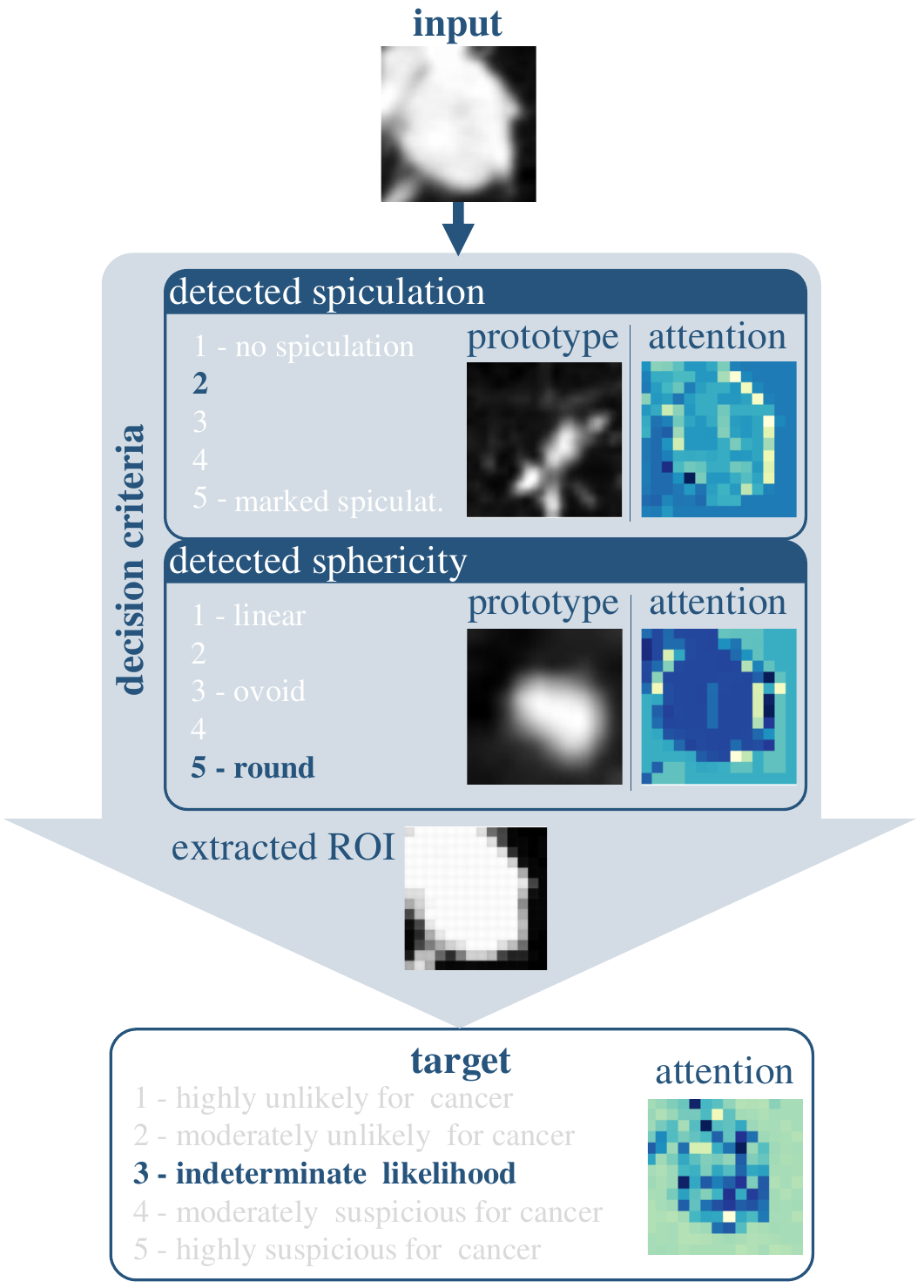}
    \caption{\textbf{HierViT} A hierarchical decision process first extracts diagnostic criteria (attributes) and the region of interest. Target prediction is based on the extracted attribute information. The attributes are defined by humans, explained by a rating, an attention heatmap, and a prototype image illustrating a similar sample.}
    \label{fig:hvitga}
\end{figure}

Our proposed model, HierViT, combines the Vision Transformer with established strategies and integrates prototype learning, domain knowledge, and a hierarchical, feature-focused prediction strategy (see Fig. \ref{fig:hvitga}). Just as radiologists rely on a structured approach to evaluate features before reaching a conclusion, HierViT mirrors this by identifying essential criteria prior to final output. This method fosters AI outputs that can be evaluated with statements like: "The AI recognized the pathological structure accurately," or "It missed essential features, indicating an unreliable prediction." 
By aligning model reasoning with human-defined criteria, HierViT may support increased user trust, as suggested by a user study of radiologists interacting with an attribute- and prototype-based AI diagnostic system \cite{gallee2024evaluating}.
In that study, radiologists diagnosed lung nodules with support from an AI that reported malignancy and attribute scores such as spiculation, margin sharpness, and calcification, each illustrated by a prototype image. Providing these feature‑level explanations increased the users' trust in the AI’s predictions. We extend this approach to Vision Transformers, leveraging their performance capabilities and built‑in attention heatmaps as an additional interpretability aid for radiologists \cite{reyes2020interpretability}.

The contributions of this work can be summarized as follows:
\begin{itemize}
    \item \textbf{Hierarchical ViT}: We propose a ViT-based model that integrates hierarchical prediction with feature-specific prototype learning for image classification.
    \item \textbf{Multifaceted interpretability}: HierViT combines predefined feature reasoning through feature scores, case-based prototypes, and attention visualizations.
    \item \textbf{State-of-the-art performance with interpretability support}: HierViT is evaluated on two medical benchmark datasets, achieving competitive performance on LIDC-IDRI and comparable performance on derm7pt, while additionally providing multiple complementary interpretability mechanisms.
\end{itemize}
The code is available at \url{https://github.com/XRad-Ulm/HierViT}.

\section{Related Work}
\paragraph{Transformer networks}
Transformer networks originated in natural language processing and were adapted for images by Dosovitskiy \textit{et al.} \cite{dosovitskiy2021vit}. Alongside Dense, Residual, and Convolutional Neural Networks, Vision Transformers represent another high-performance architecture for computer vision tasks. A Vision Transformer processes an image into patches, which are linearly embedded as patch tokens. Each token includes a positional embedding to maintain spatial information, and a learnable classification token is added for classification. The sequences are then processed by a transformer encoder using a self-attention mechanism, followed by a multi-layer perceptron (MLP) head that produces the classification result.

We adapt the standard ViT model to add interpretability by design \cite{rudin2019stop}. The target classification layer is preceded by transformer layers for learning pre-defined features, making them discriminative for target prediction. Furthermore, prototype layers are used to visualize these features.

\paragraph{Explainability by prototypes}
Prototype learning has become a widely adopted approach for interpretable image classification. Early CNN-based methods, such as ProtoPNet \cite{NEURIPS2019_adf7ee2d}, explain predictions through prototypical image regions, providing case-based explanations of the form “this looks like that”. Several subsequent works extended this idea to Vision Transformers \cite{demir2024explainable,zhu2025interpretable}: ProtoPFormer \cite{xue2922ProtoPFormer} and the more recent ProtoViT \cite{ma2024interpretable} employ ViT backbones together with spatial prototypes that represent discriminative image regions. COVID-ProtoPFormer \cite{xu2024Interpretable} demonstrated the applicability of this paradigm to medical image classification.
While these approaches provide valuable visual evidence by highlighting image regions that contribute to a prediction and linking them to prototypical image parts from the training set, the semantic meaning of the learned prototypes remains implicit. As a result, they can indicate \textit{where} the model focuses and provide a prototypical example of the corresponding image region, but not necessarily \textit{why} it supports a particular prediction, i.e., which underlying characteristic is shared between the prototype and the input image. In medical imaging, where decisions are commonly assessed against clinically established diagnostic features, this limits the ability to relate prototypes to specific diagnostic findings. HierViT addresses this challenge by incorporating domain knowledge into the prototype-learning process, such that prototypes are associated with specific, predefined clinical features, hereafter referred to as attributes.

\paragraph{Explainability by attributes}
A complementary line of research focuses on concept- or attribute-based explanations. Concept Bottleneck Models (CBMs) \cite{koh2020concept} and related approaches, such as X-Caps \cite{lalonde_encoding_2020}, introduce an intermediate layer of human-interpretable concepts, enforcing a structured prediction pipeline from attributes to the final target. These models improve semantic transparency by providing explicit attribute predictions. However, their explanations are limited to scalar concept predictions without direct visual grounding.
More recent transformer-based approaches, such as ConceptFormer \cite{rigotti2022attentionbased}, integrate concept representations into Vision Transformers via cross-attention mechanisms. While this improves flexibility and integration of concepts into modern architectures, the explanations remain at the level of concept activations and do not provide explicit visual examples of what each concept looks like in the input data.
The proposed HierViT extends this line of work by combining attribute-based reasoning with prototype-based visual grounding. In addition to predicting clinically meaningful attribute scores, HierViT retrieves attribute-specific prototypes that serve as concrete visual examples of each concept. Another advantage of HierViT in the ViT setting is the availability of attribute-specific attention maps, similar to \cite{Gamage2023melanoma, meedeniya2024skin}. By modeling attributes individually, the proposed architecture enables the extraction of self-attention maps for each attribute. Compared to attention heatmaps at the target-class level, this provides more fine-grained explanations.

Our work is most closely related to Proto-Caps \cite{gallee2023interpretable}, which combines attribute prediction with prototype-based explanations in a CNN-based architecture using Capsule Networks. HierViT extends this direction by adapting the framework to a high-performing Vision Transformer backbone, replacing convolutional feature extraction with transformer-based representations, while aligning attribute prediction and prototype retrieval within a shared hierarchical structure.
This design produces three complementary components that support the target disease classification: attribute-level predictions, attribute-specific prototypes, and attention maps derived from ViT self-attention and associated with individual attributes. In particular, attribute-specific prototypes provide visual exemplars for the predicted attribute scores. While the employed attributes correspond to established clinical concepts, their assessment is often subjective and may exhibit substantial inter-rater variability, especially for ordinal scales such as those used in lung nodule malignancy assessment (e.g., sphericity, which ranges from 1-\textit{linear} to 5-\textit{round}) \cite{zhang2022re}. By linking predicted scores to representative examples from the training data, the prototypes make the model's interpretation of these concepts more explicit and provide case-based evidence for its predictions. Attribute-specific attention maps complement the prototype-based explanations by indicating where in the image the evidence supporting the predicted attribute score and associated prototype is located. Together, these elements provide both semantic and case-based explanations within a single model.

\section{Method}
\begin{figure*}[!htbp]
    \centering
    \includegraphics[width=0.6\linewidth]{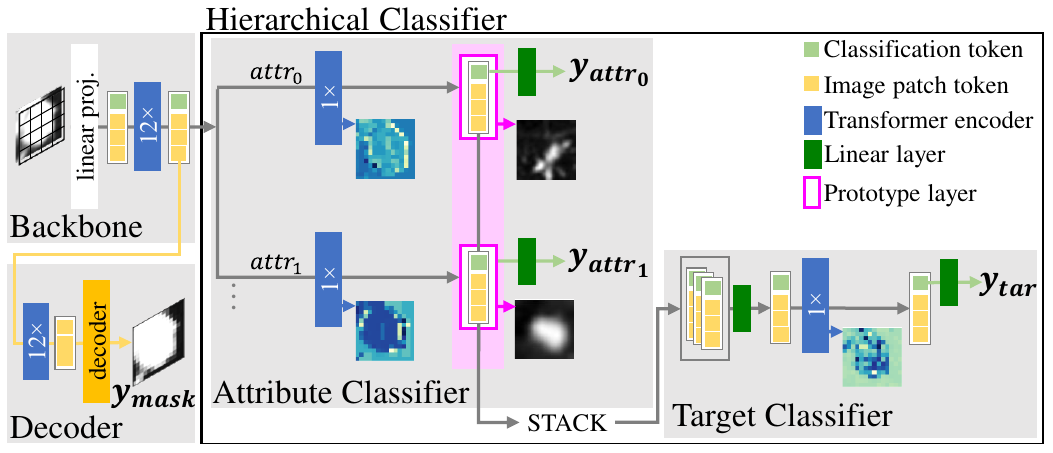}
    \caption{\textbf{Proposed model} The patchified image is linearly projected and processed by a transformer encoder, producing a token vector that serves as the input for both a decoder and a hierarchical classifier. The optional decoder segments a region-of-interest mask. The hierarchical classifier processes the token vector through multiple transformer layers, one for each attribute, with individual heads providing attribute ratings. For target prediction, the token vectors from the attribute layers are stacked and further processed by the target branch.}
    \label{fig:modelarchitecture}
\end{figure*}

The proposed model uses a ViT-based backbone as an encoder, featuring two branches derived from the extracted features (see Fig. \ref{fig:modelarchitecture}). The first branch is an optional ViT-based decoder for creating a segmentation mask if labels are available. The second branch functions as a hierarchical classifier, mapping extracted features to predefined attributes for target classification. Additionally, a prototype layer enables case-based reasoning at the attribute level.
The information process of HierViT can be summarised to $X \rightarrow Y$, with $X$ denoting the input image, and $Y:\{y_{tar},\vec{y}_{attr},y_{mask},\vec{y}_{proto}\}$ denoting the model prediction of the target $y_{tar}$, attributes $\vec{y}_{attr}$, segmentation $y_{mask}$, and attribute prototypes $\vec{y}_{proto}$.

\subsection{Architecture}
\paragraph{Backbone}
The ViT encoder is implemented as the original Vision Transformer architecture described by Dosovitskiy \textit{et al.} \cite{dosovitskiy2021vit}, enabling initialization with pretrained weights. The encoder operates on input images of size 224\,×\,224 pixels, which are divided into 16\,×\,16 patches. It utilizes twelve encoder layers, and the resulting token vector serves as the foundation for all subsequent prediction paths of the model.

\paragraph{Hierarchical classifier}
The predictions rely on the output token extracted from the ViT-based backbone. First, individual high-level attributes are learned, which are human-defined features describing the object's visual characteristics.
Each attribute score is computed using a dedicated one-layer transformer encoder and a linear prediction head. Specifically, the classification token from the encoder output serves as input to the linear layer, which predicts the corresponding attribute score.
The self-attention maps of the corresponding transformer encoder additionally serve as attribute-specific attention heatmaps.
The loss function term for attribute learning $\mathcal{L}_{attr}$ minimizes the mean value of the classification error $\mathcal{L}_{class}$ over all attributes. 
In the following $y_{attr_a}$ is the ground truth label of attribute $a=1...A$, and $\hat{y}_{attr_a}$ is the respective prediction:
\begin{equation}
\mathcal{L}_{attr} = \frac{1}{A} \sum^A_a \mathcal{L}_{class}(y_{attr_a},\hat{y}_{attr_a}).
\end{equation}

Second, the target prediction is based on the encoded attribute information. The outputs of the attribute-specific encoders are stacked and fed into the target classifier. A linear layer combines these features into a single token vector, which is then processed by a target transformer encoder layer. Finally, classification is performed by a linear layer on the resulting classification token.
The target loss function $\mathcal{L}_{tar}$ represents the classification error between the ground truth $y_{tar}$ and the predicted value $\hat{y}_{tar}$:
\begin{equation}
\mathcal{L}_{tar} = \mathcal{L}_{class}(y_{tar},\hat{y}_{tar}).
\end{equation}

Depending on the scale of the data, either the mean square error (MSE) or the cross entropy loss (CSE) was chosen for the classification loss function $\mathcal{L}_{class}$:

\begin{equation}
\mathcal{L}_{class} = 
\begin{cases} 
\text{MSE}(y, \hat{y}) & \text{for ordinal data (LIDC-IDRI),} \\ 
\text{CSE}(y, \hat{y}) & \text{for nominal data (derm7pt).}
\end{cases}
\end{equation}

\paragraph{Prototype layer}
Prototypes serve as visual examples of the extracted attributes. Unlike most prototype-based networks that generate prototypes at the target level, our method focuses on attribute-specific prototypes. In medical image diagnosis, classification relies on the variance of critical visual criteria, which can take many forms. Using attribute-specific prototypes enables the representation of detailed features with greater granularity than a single target-level prototype allows.

In the proposed method, the prototypes are derived from the attribute vectors. Each prototype layer consists of a set of learnable vectors representing different attribute values, thereby enabling the mapping of diverse characteristics within the same attribute value. 
The loss function $\mathcal{L}_{proto}$ encourages training samples to be similar to the prototypes of the correct attribute class. 
It is implemented by the Euclidean distance between the sample's attribute vector $\vec{c}^{\,a}$ and the prototypes $\vec{p}^{\,a,p}$, where $a$ denotes the respective attribute, and $p = 1,\dots,P$ indexes the prototypes within the ground-truth class of attribute $a$.
The set of prototypes corresponding to the correct class for attribute $a$ is defined as $\mathcal{P}_a = \{ \vec{p}^{\,a,p} \}_{p=1}^{P}$.
\begin{equation}\label{eq:L_proto}
    \mathcal{L}_{proto} = \frac{1}{A} \frac{1}{P} \sum^A_{a=1} \sum_{p=1}^P \left\Vert \vec{c}^{\,a}-\vec{p}^{\,a,p}\right\Vert_2 .
\end{equation}

A push operation assigns to each prototype vector the training sample whose attribute representation is closest in feature space. Specifically, for each prototype, the Euclidean distance to all training samples is computed, and the sample with the minimum distance is selected. The selected samples serve as visual representatives of the prototypes, enabling visualization with real images, as shown in Section~\ref{subsec:explainability}. 
The frequency of the push operation is controlled by the hyperparameter \textit{push step}.
Between push operations, the prototype vectors continue to be optimized using the current attribute representations of the training samples and the loss $\mathcal{L}_{proto}$.

\paragraph{Decoder}
The decoder branch computes the segmentation mask and can be optionally enabled if segmentation masks are available in a dataset, allowing the model to leverage additional spatial supervision and focus on the region of interest during training.
Symmetrically to the encoder, twelve transformer layers process the image tokens of the ViT backbone. To restore the image size from the tokenised vector, it is first reshaped from the flattened dimension to a two-dimensional transposed convolution, then scaled to pixel size, followed by batch normalization, sigmoid activation, and averaging along the hidden dimension to return the original image size.

The segmentation loss term $\mathcal{L}_{seg}$ calculates the mean square error between the segmentation mask label $y_{mask}$ and the decoder prediction $\hat{y}_{mask}$:
\begin{equation}
    \mathcal{L}_{seg} =  \text{MSE}(y_{mask}, \hat{y}_{mask})
\end{equation}
MSE is used to align the optimization characteristics of the segmentation loss with the classification losses, thereby supporting stable joint training.

\subsection{Training Algorithm}
\label{subsec:trainingalgoLOSS}
The model can simultaneously address several semantically related tasks, including object region segmentation, extraction of specific high-level visual features, target prediction based on these features, and generation of attribute-specific prototypes. Previous work \cite{gallee2023interpretable} shows that prototype learning should begin only after a warm-up phase, during which the model weights are adjusted to the core task. The loss function is therefore composed as follows for the warm-up phase: 
\begin{equation}
    \mathcal{L}_{warm-up} = \mathcal{L}_{tar}+\mathcal{L}_{attr}+\mathcal{L}_{seg}, 
\end{equation}
and for the final phase:
\begin{equation}
    \mathcal{L}_{final} = \mathcal{L}_{tar}+\mathcal{L}_{attr}+\mathcal{L}_{seg}+\lambda_{proto}\cdot\mathcal{L}_{proto}.
    \label{eq:combined_loss_function}
\end{equation}

The hyperparameter $\lambda_{proto}$ was set to $0.01$ to ensure the model remained focused on the primary task. A detailed analysis of this hyperparameter is presented in Section~\ref{subsec:hyperparameter_analysis}.

\section{Datasets}

\subsection{LIDC-IDRI}
The Lung Image Database Consortium and Image Database Resource Initiative (CC BY 3.0 licensed) \cite{armato_iii_lung_data_2015} is a large, extensively annotated medical dataset of chest CT scans from patients with non-small cell lung cancer. Up to four radiologists segmented lung nodules and labeled their appearance and malignancy \cite{armato_iii_lung_2011}. In our experiments, we used lung nodule cropout images as input, segmentation masks as decoder targets, malignancy ratings as prediction targets, and appearance ratings as attributes, see Table \ref{tab:lidcidridataset}.
Each radiologist's annotation was treated as an individual sample, using the corresponding radiologist's segmentation mask as the decoder target, while malignancy and appearance ratings were obtained by averaging the scores across all available radiologists for the respective nodule.

\begin{table}[ht]
    \centering
    \setlength{\tabcolsep}{4pt}
    \begin{tabular}{llllll}
         \toprule
         \multicolumn{2}{c}{\textbf{nodule image}} & \multicolumn{2}{c}{\textbf{nodule mask}} & \multicolumn{2}{c}{\textbf{target: malignancy}} \\
         \multicolumn{2}{c}{
         \begin{minipage}{.15\linewidth}
            \includegraphics[width=\linewidth,]{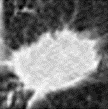}
         \end{minipage}}
         &\multicolumn{2}{c}{
         \begin{minipage}{.15\linewidth}
            \includegraphics[width=\linewidth]{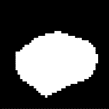}
         \end{minipage}}
         &\multicolumn{2}{l}{
         \makecell{1-\textit{highly unlikely},\\5-\textit{highly suspicious}}}\\
         \midrule
         \multicolumn{6}{l}{\textbf{attributes}} \\
         \multicolumn{3}{l}{$\bullet$ subtlety 1-\textit{extremely sub}, 5-\textit{obvious}}
         &
         \multicolumn{3}{l}{$\bullet$ margin 1-\textit{poorly def}, 5-\textit{sharp}}\\
         \multicolumn{3}{l}{$\bullet$ internal structure 1-\textit{soft tissue}, 4-\textit{air}}
         &
         \multicolumn{3}{l}{$\bullet$ lobulation 1-\textit{none}, 5-\textit{marked}}\\
         \multicolumn{3}{l}{$\bullet$ calcification 1-\textit{popcorn}, 6-\textit{absent}}
         &
         \multicolumn{3}{l}{$\bullet$ spiculation 1-\textit{none}, 5-\textit{marked}}\\
         \multicolumn{3}{l}{$\bullet$ sphericity 1-\textit{linear}, 5-\textit{round}}
         &
         \multicolumn{3}{l}{$\bullet$ texture 1-\textit{non-solid}, 5-\textit{solid}}\\         
         \bottomrule
    \end{tabular}
    \caption{Annotations of the LIDC-IDRI dataset used in experiments.}
    \label{tab:lidcidridataset}
\end{table}

Data preprocessing excluded nodules identified by fewer than three radiologists or smaller than 3\,mm. Cropouts were created using the smallest square bounding box for the nodules, and each slice was resized to 224 × 224 pixels, implemented with the \texttt{pylidc} framework \cite{LIDC_hancock}. The final dataset includes 27,379 samples and was evaluated using 5-fold stratified cross-validation by patient, with $10\,\%$ of the training data used for validation.

\subsection{derm7pt}
The derm7pt dataset is a publicly available medical benchmark for dermatology, containing comprehensive annotated images of skin lesions \cite{kawahara2019seven}. It includes 1,011 samples, each consisting of a clinical and a dermoscopic image, along with metadata such as the patient’s gender and the anatomical location of the skin lesion. 
The lesion annotations include diagnosis labels and seven attributes describing the visual appearance of lesions, which are commonly used by dermatologists \cite{argenziano1998epiluminescence}.
In this study, we use dermoscopic images as input data, as they provide a standardized view with fewer artifacts compared to clinical images. Lesion classification targets include nevus, seborrheic keratosis, miscellaneous, basal cell carcinoma, and melanoma. Visual attributes, such as pigment network, blue whitish veil, vascular structures, pigmentation, streaks, dots and globules, and regression structures, are detailed in Table \ref{tab:derm7ptdataset} based on the work of Kawahara \textit{et al.} \cite{kawahara2019seven}. Unlike the LIDC-IDRI dataset, derm7pt does not include segmentation masks for regions of interest.

\begin{table}[ht]
    \centering
    \begin{tabular}{cc}
         \toprule
         \textbf{lesion image}&\textbf{target: diagnosis}\\
         \begin{minipage}{.15\linewidth}
            \includegraphics[width=\linewidth,]{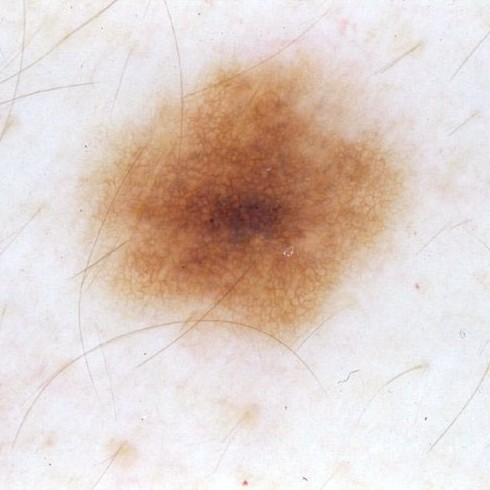}
         \end{minipage}
         &
         \begin{minipage}{.75\linewidth}
            \centering
            \textit{nevus (nev),}\\
            \textit{seborrheic keratosis (sk),}\\
            \textit{miscellaneous (misc),}\\
            \textit{basal cell carcinoma (bcc),}\\
            \textit{melanoma (mel)}
         \end{minipage}
         \\
         \midrule
         \multicolumn{2}{l}{\textbf{attributes}} \\
         \multicolumn{2}{l}{$\bullet$ pigment network (pn): \textit{absent, typical, atypical}}\\
         \multicolumn{2}{l}{$\bullet$ blue whitish veil (bmw): \textit{absent, present}}\\
         \multicolumn{2}{l}{$\bullet$ vascular structures (vs): \textit{absent, regular, irregular}}\\
         \multicolumn{2}{l}{$\bullet$ pigmentation (pig): \textit{absent, regular, irregular}}\\
         \multicolumn{2}{l}{$\bullet$ streaks (str): \textit{absent, regular, irregular}}\\
         \multicolumn{2}{l}{$\bullet$ dots and globules (dag): \textit{absent, regular, irregular}}\\
         \multicolumn{2}{l}{$\bullet$ regression structures (rs): \textit{absent, present}}\\       
         \bottomrule
    \end{tabular}
    \caption{Annotations of the derm7pt dataset used in experiments.}
    \label{tab:derm7ptdataset}
\end{table}

Data preprocessing included centered cropping to 450 × 450 pixels and resizing to 224 × 224 pixels. Training samples were randomly rotated horizontally and vertically. For comparability with related work, we used the test split from Kawahara \textit{et al.} \cite{kawahara2019seven}.

\section{Experiments and Results}

Since the proposed method aims to be an interpretable medical assistance tool, the evaluation experiments cover multiple aspects. Alongside a quantitative performance comparison with current SOTA methods, the explainability aspect is analyzed qualitatively and through a detailed analysis of the prototypes, which contributes to the method's explainability. Additionally, various ablation studies on the model and annotation availability assess functionality. We further examine the impact of user interaction with the model through interventions on incorrect attribute detections.

\subsection{Experiment setting} \label{subsec:experimentsetting}
For the experiments, we chose the ViT-Base model by Dosovitskiy \textit{et al.} \cite{dosovitskiy2021vit} as backbone and weights pre-trained on ImageNet-1K.
A total of $P=16$ prototypes were made available for each attribute class (LIDC-IDRI: $A=8$, derm7pt: $A=7$), following prior work on prototype-based representation learning \cite{gallee2023interpretable}. In order to reduce the training time, the push operation was repeated every other epoch by setting the parameter \textit{push step=2}.
\paragraph{LIDC-IDRI}
The model parameters were optimized using an Adam optimizer with a learning rate of $0.001$. 
A warm-up phase of two epochs was chosen for prototype learning once the validation accuracy was sufficient. 

The LIDC-IDRI experiments on a GeForce RTX 3090 graphics card, conducted over 30 epochs, yielded an average training time of 18 hours.
The model comprises 484.6M trainable parameters in total, distributed across the main components shown in Fig. \ref{fig:modelarchitecture}: 86.6M Backbone, 57.9M Attribute Classifier, 96.8M Prototypes, 7.2M Target Classifier, and 236.1M Decoder.

\paragraph{derm7pt}
The model layers were optimized using an Adam optimizer. Preliminary tests were used to determine the optimal learning rate of $0.00001$ for all layers except the prototype vectors, which were optimized with a learning rate of $0.01$. A warm-up phase of twenty epochs was chosen. 
As the derm7pt dataset does not provide segmentation masks, the decoder branch was deactivated by setting the segmentation loss function $L_{seg}=0$. To compensate for the imbalance in the data set, appropriate weighting of the target and attribute classes was implemented in the cross-entropy loss terms.

The derm7pt experiments on a GeForce RTX 3090 graphics card, conducted over 400 epochs, yielded an average runtime of four hours.
The model comprises 190.5M trainable parameters in total, fewer than the LIDC-IDRI model due to the absence of the decoder and the smaller number of attributes and classes: 86.6M Backbone, 50.7M Attribute Classifier, 46.0M Prototypes, 7.2M Target Classifier.

\subsection{Qualitative Evaluation}
\label{subsec:explainability}
The hierarchical architecture of the model enables the generation of prediction results that build upon one another. High-level visual attributes are first learned from the latent vector of the ViT encoder, and target prediction follows based on this information. This step-by-step flow mimics human reasoning to facilitate a comparable thought process and discussion of the results.
\begin{figure*}[!htbp]
    \centering
    \includegraphics[width=0.8\linewidth]{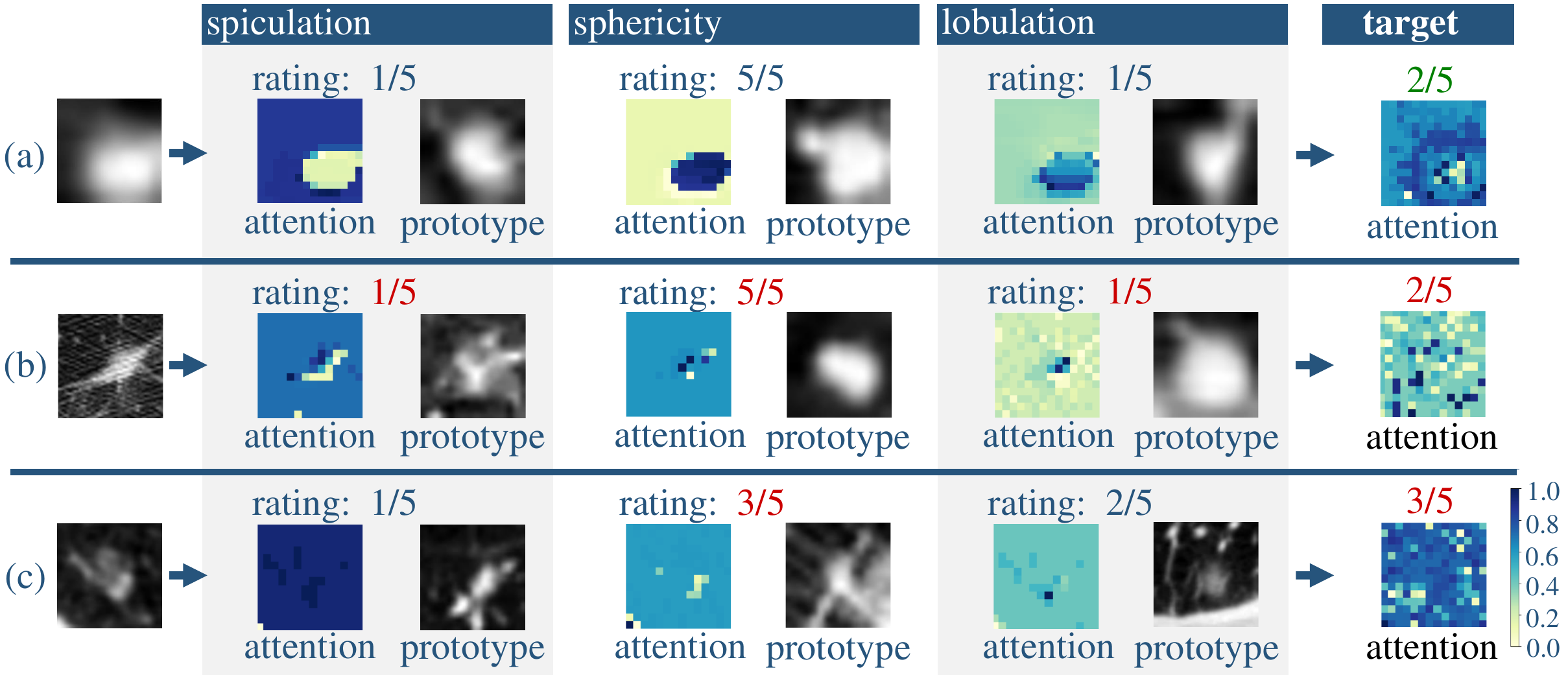}
    \caption{\textbf{Reasoning process} Three sample cases are illustrated, (a) correctly predicted, (b) and (c) incorrectly predicted. For three of the eight attributes (spiculation, sphericity, lobulation), the score, attention heatmap, and prototype image of the respective attribute are displayed.
    }
    \label{fig:hierarchicalreasoning}
\end{figure*}

The model output includes three predictions of the detected attributes that justify the target prediction: associated scores, attention heatmaps, and the closest prototypical samples. Fig. \ref{fig:hierarchicalreasoning} illustrates the model output, showcasing three of the eight attributes.
In case (a), the model correctly predicted the target. The attribute ratings reflect the visual characteristics of the sample nodule, and the closest attribute prototypes exhibit similar traits. The attention heatmaps further support the model's prediction by highlighting the attribute-specific region of interest. Consultation with a pulmonary nodules expert confirmed the significance of the attention areas, with the model focusing on the nodules' edges for spiculation assessment and their interior for evaluating sphericity and lobulation.

In cases (b) and (c), the attention heatmaps and prototypes indicate a misclassification, showing a discrepancy between the prototypes' characteristics and their ratings compared to the inference image for certain attributes. Additionally, the attention is not focused on the nodule. These signals should raise user doubts about the model's results, helping prevent incorrect conclusions in the diagnosis.

Compared to 3D-CNN+MTL \cite{hussein_risk_2017} and TumorNet \cite{hussein_tumornet_2017}, HierViT provides interpretability by predicting attributes as domain knowledge. 
While X-Caps \cite{lalonde_encoding_2020} and Concept Bottleneck \cite{koh2020concept} focus on predicting attribute values, Proto-Caps \cite{gallee2023interpretable} and our proposed method, HierViT, extend this approach by also providing prototypical visualizations of the corresponding attribute characteristics. Unlike Proto-Caps, which employs an iterative attention mechanism, HierViT leverages Vision Transformer (ViT) layers. This design choice allows for the extraction and visualization of attention regions associated with attribute predictions, thereby enhancing model interpretability, as illustrated in Fig.~\ref{fig:hierarchicalreasoning}.

\subsection{Quantitative Evaluation}
\paragraph{LIDC-IDRI}
In accordance with previous studies on the LIDC-IDRI dataset \cite{gallee2023interpretable, lalonde_encoding_2020}, the proposed model was evaluated using the Within-1-Accuracy metric. This metric for ordinal labels defines a prediction with a tolerance of one point value to the ground truth label as correct. 

All results were taken from the respective publications, except for Concept Bottleneck \cite{koh2020concept}, which we evaluated on the LIDC-IDRI dataset ourselves, as it was not included in the original paper. For the Concept Bottleneck experiments, we adopted the default settings of the 'joint' configuration, using a model with four fully connected layers of output sizes \texttt{[256, 128, 8, 5]}. The third fully connected layer predicts the eight visual attributes, while the final layer outputs classification scores for the five target classes.

Although Concept Bottleneck achieves the highest mean performance on the individual attributes margin and spiculation, our proposed method, HierViT, achieves the highest mean performance averaged across all attributes and the highest mean target classification performance among the compared methods. The $95\%$ binomial confidence intervals for target classification do not overlap between HierViT and the competing methods, as shown in Table~\ref{lidcPerformance}.

\begin{table*}[ht]
  \caption{\textbf{Performance comparison LIDC-IDRI} The performance of the experiments on the LIDC-IDRI dataset is reported in the Within-1-Accuracy metric (for ordinal rating 1-5). In contrast, the results marked with an asterisk * represent the accuracy (ACC) of the binary classification between benign and malignant cases.  
  Mean (black) and standard deviation (grey) are calculated from 5-fold experiments and reported if available in the literature. Methods denoted with "P" provide reasoning with prototypes. The $95\%$ binomial confidence interval is reported for the target if the test dataset is detailed in the literature. The highest value is in bold.}
  \label{lidcPerformance}
  \centering
  \begin{tabular}{lcccccccccl}
    \toprule
    &&\multicolumn{8}{c}{attributes} & \multicolumn{1}{c}{target}\\
    \cmidrule(lr{0.5em}){3-10}
    \cmidrule(lr{0.5em}){11-11}
    && sub & is & cal & sph & mar & lob & spic & tex & \multicolumn{1}{c}{malignancy}\\
    \midrule
    \textbf{CNN-based}&&&&&&&&&&\\
    3D-CNN+MTL \cite{hussein_risk_2017}&&-&-&-&-&-&-&-&-&91.3 [89.7,92.9]\\
    TumorNet \cite{hussein_tumornet_2017}&&-&-&-&-&-&-&-&-&92.3 [90.8,93.8]\\
    X-Caps \cite{lalonde_encoding_2020}&&90.4&-&-&85.4&84.1&70.7&75.2&93.1&86.4\\
    Concept Bottleneck \cite{koh2020concept}&&94.7&99.0&93.9&97.0&\textbf{94.6}&89.0&\textbf{92.4}&92.6&91.6 [91.3,91.9]\\
    &&\textcolor{gray}{1.5}&\textcolor{gray}{1.5}&\textcolor{gray}{2.3}&\textcolor{gray}{2.5}&\textcolor{gray}{0.9}&\textcolor{gray}{2.3}&\textcolor{gray}{3.8}&\textcolor{gray}{1.0}&\textcolor{gray}{1.6}\\
    Proto-Caps \cite{gallee2023interpretable}&P&89.1&\textbf{99.8}&95.4&96.0&88.3&87.9&89.1&\textbf{93.3}&93.0 [92.7,93.3]\\
    &&\textcolor{gray}{5.2}&\textcolor{gray}{0.2}&\textcolor{gray}{1.3}&\textcolor{gray}{2.2}&\textcolor{gray}{3.1}&\textcolor{gray}{0.8}&\textcolor{gray}{1.3}&\textcolor{gray}{1.0}&\textcolor{gray}{1.5}\\
    \cmidrule{1-1}
    \textbf{ViT-based}&&&&&&&&&&\\
    TransUnet \cite{wang2022accurate}&&-&-&-&-&-&-&-&-&84.62*\\
    Res-trans \cite{liu2022res}&&-&-&-&-&-&-&-&-&92.92*\\
    TransPND \cite{wang2022transpnd}&&-&-&-&-&-&-&-&-&93.33*\\
    HierViT (proposed)&P&\textbf{96.3}&\textbf{99.8}&\textbf{95.5}&\textbf{97.4}&92.7&\textbf{94.3}&90.8&\textbf{93.3}&\textbf{94.8} [94.5,95.1]\\
    &&\textcolor{gray}{0.9}&\textcolor{gray}{0.3}&\textcolor{gray}{2.1}&\textcolor{gray}{0.8}&\textcolor{gray}{1.7}&\textcolor{gray}{2.9}&\textcolor{gray}{1.4}&\textcolor{gray}{1.4}&\textcolor{gray}{1.4}\\
    \bottomrule
  \end{tabular}
\end{table*}

Related research using Vision Transformers \cite{wang2022accurate, liu2022res, wang2022transpnd} focuses on the binary classification of lung nodules, summarizing malignancy annotations into benign and malignant while disregarding nodules of intermediate suspicion. These works do not prioritize model interpretability, nor do they include domain knowledge of visual appearance as attribute labels in training or predictions. Due to differences in label definitions and evaluation metrics, direct quantitative comparisons are limited. However, the reported results are included for completeness to contextualize prior work.

The proposed model HierViT achieved a Dice score of $68.2\,\%$, with a standard deviation of $2.5\,\%$ in the reconstruction of the segmentation mask.
This result indicates that the model is able to localize the region of interest at a coarse level, capturing the general object location. 

\paragraph{derm7pt}
The derm7pt dataset provides clinical images, dermoscopic images, and metadata. Since dermoscopic images are more standardized than clinical images, our method uses only the dermoscopic modality. For a fair comparison, Table~\ref{dermacc} is divided into two parts: the upper half compares methods that use only dermoscopic images, while the lower half presents results from methods that additionally incorporate clinical images and/or metadata.

Comparison results were taken from the respective publications, except for Concept Bottleneck \cite{koh2020concept}, which we evaluated on the derm7pt dataset ourselves. We used the 'joint' configuration with layer output sizes \texttt{[256, 128, 19, 5]}, adapted to the attribute and target classification tasks.

\begin{table*}[ht]
  \caption{\textbf{Performance comparison derm7pt} The performance results (in accuracy percentage) are listed, differentiated by data modality. The average $\varnothing$ is the mean over the attributes and target (diag). The $95\%$ binomial confidence interval is specified for the target and the average.
  In the HierViT's category using dermoscopic images only, the highest values are in bold, and the second-highest are underlined.}
  \label{dermacc}
  \centering
  \begin{tabular}{lcccccccrr}
    \toprule
    &\multicolumn{7}{c}{attributes}&\makecell[c]{target}&\makecell[c]{\multirow{3}{*}{$\varnothing$}}\\
    \cmidrule(lr{0.5em}){2-8}
    \cmidrule(lr{0.5em}){9-9}
    &pn&bmv&vs&pig&str&dag&rs&\makecell[c]{diagnosis}&\\
    \midrule
    \makecell[l]{\textbf{using dermoscopic}\\\textbf{images only}}&&&&&&&&&\\
    Inception-$x_d$ \cite{kawahara2019seven}&\textbf{69.4}&85.8&80.3&62.8&71.4&\textbf{60.8}&77.5&71.9 [67.5,76.3]&72.5 [68.1,76.9]\\
    AMFAM
    \cite{wang2022adversarial}&66.1&87.1&\underline{80.5}&66.6&71.1&\underline{60.0}&78.5&69.4 [64.9,73.9]&72.4 [68.0,76.8]\\
    FusionM4Net
    \cite{tang2022Fusion}&\underline{69.0}&\underline{87.2}&\textbf{81.4}&\textbf{68.3}&\underline{73.7}&\underline{60.0}&\textbf{80.1}&\underline{74.7} [70.4,79.0]&\textbf{74.3} [70.0,78.6]\\
    MTL-standard \cite{coppola2020interpreting}&55.4&85.1&63.5&62.5&49.1&48.6&65.1&45.8 [40.9,50.7]&59.4 [54.6,64.2]\\
    Concept Bottleneck \cite{koh2020concept}&67.1&85.8&80.3&\underline{67.8}&72.4&58.5&77.0&70.4 [65.9,74.9]&72.4 [68.0,76.8]\\
    HierViT (proposed)&65.3&\textbf{87.6}&80.3&67.3&\textbf{74.4}&59.2&\underline{79.8}&\textbf{76.5} [72.3,80.7]&\underline{73.8} [69.5,78.1]\\
    \cmidrule{1-1}
    \makecell[l]{\textcolor{gray}{\textbf{using dermoscopic}}\\\textcolor{gray}{\textbf{and clinical images,}}\\\textcolor{gray}{\textbf{and metadata*}}}&&&&&&&&&\\
    \textcolor{gray}{AMFAM \cite{wang2022adversarial}}&\textcolor{gray}{70.6}&\textcolor{gray}{88.1}&\textcolor{gray}{83.3}&\textcolor{gray}{70.9}&\textcolor{gray}{74.7}&\textcolor{gray}{63.8}&\textcolor{gray}{80.8}&\textcolor{gray}{75.4 [71.2,79.6]}&\textcolor{gray}{76.0 [71.8,80.2]}\\
    \textcolor{gray}{HcCCN \cite{bi2020multi}}&\textcolor{gray}{70.6}&\textcolor{gray}{87.1}&\textcolor{gray}{84.8}&\textcolor{gray}{68.6}&\textcolor{gray}{71.6}&\textcolor{gray}{65.6}&\textcolor{gray}{80.8}&\textcolor{gray}{69.9 [65.4,74.4]}&\textcolor{gray}{74.9 [70.6,79.2]}\\
    \textcolor{gray}{Inception-combine* \cite{kawahara2019seven}}&\textcolor{gray}{70.9}&\textcolor{gray}{87.1}&\textcolor{gray}{79.7}&\textcolor{gray}{66.1}&\textcolor{gray}{74.2}&\textcolor{gray}{60.0}&\textcolor{gray}{77.2}&\textcolor{gray}{74.2 [69.9,78.5]}&\textcolor{gray}{73.7 [69.4,78.0]}\\
    \textcolor{gray}{FusionM4Net* \cite{tang2022Fusion}}&\textcolor{gray}{71.1}&\textcolor{gray}{88.1}&\textcolor{gray}{81.8}&\textcolor{gray}{70.1}&\textcolor{gray}{78.0}&\textcolor{gray}{66.1}&\textcolor{gray}{81.5}&\textcolor{gray}{78.5 [74.4,82.6]}&\textcolor{gray}{77.0 [72.8,81.2]}\\
    \bottomrule
  \end{tabular}
\end{table*}

The HierViT method achieves comparable mean accuracy to related works using only dermoscopic images, with mean target prediction accuracy and average attribute and target accuracy ranking highest and second-highest, respectively, although the $95\%$ binomial confidence intervals substantially overlap with those of FusionM4Net.

Detailed performance results, including sensitivity, specificity, precision, AUC, and ROC curves, are provided in Appendix Table \ref{dermtarget} and \ref{dermattr} and Figure \ref{fig:rocs}.

Any method that accurately predicts the seven lesion features provides a degree of interpretability. The Inception model \cite{kawahara2019seven} and the AMFAM model \cite{wang2022adversarial} further enhance interpretability through visualization of prediction importance heatmaps. HierViT also captures the hierarchical relationship between lesion features and classification in the derm7pt dataset, treating it as more than a multi-label task.

\subsection{Ablation study}
The ablation studies on the LIDC-IDRI dataset were conducted with the objective of evaluating the efficacy of the individual components of the model architecture. Four reduced variants were subjected to comparative analysis with the proposed method, see Table \ref{lidcablation}.
\begin{table*}[ht]
  \caption{\textbf{Ablation study} Performance of the proposed method when model components were abstracted. The experiments were conducted with the LIDC-IDRI dataset, and the Within-1-Accuracy was measured. Mean (in black) and standard deviation (in grey) are calculated from 5-fold experiments. The $95\%$ binomial confidence interval is reported for the target.}
  \label{lidcablation}
  \centering
  \begin{tabular}{lccccccccl}
    \toprule
    &\multicolumn{8}{c}{attributes} &\multicolumn{1}{c}{target}\\
    \cmidrule(lr{0.5em}){2-9}
    \cmidrule(lr{0.5em}){10-10}
    & sub & is & cal & sph & mar & lob & spic & tex &\multicolumn{1}{c}{malignancy}\\
    \midrule
    HierViT (proposed)&96.3&99.8&95.5&97.4&92.7&94.3&90.8&93.3&94.8 [94.5,95.1]\\
    &\textcolor{gray}{0.9}&\textcolor{gray}{0.3}&\textcolor{gray}{2.1}&\textcolor{gray}{0.8}&\textcolor{gray}{1.7}&\textcolor{gray}{2.9}&\textcolor{gray}{1.4}&\textcolor{gray}{1.4}&\textcolor{gray}{1.4}\\
    A. $+$ proto. inference&93.7&99.8&95.1&92.2&87.9&88.7&86.0&93.0&94.4 [94.1,94.7]\\
    &\textcolor{gray}{2.0}&\textcolor{gray}{0.3}&\textcolor{gray}{1.8}&\textcolor{gray}{7.5}&\textcolor{gray}{3.4}&\textcolor{gray}{3.7}&\textcolor{gray}{1.9}&\textcolor{gray}{3.2}&\textcolor{gray}{1.9}\\
    B. $-$ proto. learning&95.6&99.8&95.9&97.3&91.5&94.3&91.3&93.9&95.0 [94.7,95.3]\\
    &\textcolor{gray}{1.7}&\textcolor{gray}{0.3}&\textcolor{gray}{1.7}&\textcolor{gray}{0.7}&\textcolor{gray}{2.8}&\textcolor{gray}{2.1}&\textcolor{gray}{1.9}&\textcolor{gray}{1.8}&\textcolor{gray}{2.0}\\
    C. $-$ attribute learning&-&-&-&-&-&-&-&-&93.7 [93.4,94.0]\\
    &-&-&-&-&-&-&-&-&\textcolor{gray}{2.5}\\
    D. $-$ segmentation&96.5&99.8&96.0&97.6&92.5&94.5&91.3&93.5&95.6 [95.4,95.8]\\
    &\textcolor{gray}{1.2}&\textcolor{gray}{0.3}&\textcolor{gray}{1.9}&\textcolor{gray}{0.7}&\textcolor{gray}{1.6}&\textcolor{gray}{1.9}&\textcolor{gray}{1.2}&\textcolor{gray}{1.8}&\textcolor{gray}{1.2}\\
    \bottomrule
  \end{tabular}
\end{table*}

\paragraph{A. Using prototypes during inference}
\label{para:A.}
This experiment follows the same training procedure as the proposed method. 
However, during inference, the token vectors of the attributes are replaced by the closest prototype vectors for target prediction, similar to Proto-Caps \cite{gallee2023interpretable}. 
The exemplar sample linked to the prototype vector is used for predicting attribute scores based on its ground truth label, while the attribute heads are ignored. 
Although this variant reduces the mean Within-1-Accuracy values for certain attributes, it enforces a more direct use of prototypes during inference, thereby strengthening their role in the prediction process and improving case-based interpretability.

\paragraph{B. Prototype learning}
In this experiment, the prototype layer was deactivated, reducing the explainability of target prediction to attribute scores and attention outputs. 
The results show an overlap in the $95\%$ binomial confidence interval between this reduced version and the original method, suggesting that prototype learning is not necessary for achieving comparable target prediction performance. However, its contribution lies primarily in providing an additional, case-based explanation modality: while attribute scores describe predefined clinical characteristics, prototypes provide visual reference points for these characteristics. Thus, prototype learning complements the other explanation modalities by providing additional visual evidence without negatively impacting predictive performance.

\paragraph{C. Attribute learning}
This variant aims to assess the impact of incorporating domain knowledge through attribute labels on target performance. The model was optimized only for the target label and segmentation mask, not for attributes. 
The observed decline of 1.1\% in target prediction performance, measured by Within-1-Accuracy, suggests that attribute supervision may provide a beneficial inductive bias toward clinically meaningful intermediate representations, potentially improving generalization compared to optimizing solely for the target label.
In addition, attribute prediction improves the interpretability of the model by introducing clinically meaningful intermediate concepts.

\paragraph{D. Segmentation learning}
This experiment simulates the lack of segmentation mask annotations. Deactivating the decoder branch yields comparable performance in attribute prediction and an increase of 0.8\% Within-1-Accuracy for target prediction compared to the complete model, with non-overlapping confidence intervals.
The segmentation branch is not a functional requirement of the hierarchical classification framework and can be omitted in settings where segmentation masks are unavailable, while also reducing the number of model parameters, see Section \ref{subsec:experimentsetting}. However, when region-of-interest annotations are provided, the segmentation branch enables the use of additional spatial supervision and can support interpretability by explicitly modeling the region of interest.

\subsection{Test-time intervention}
HierViT's architecture enables interaction at the attribute level, similar to \cite{zhang2024decoupling}. In a radiology setting, this allows the radiologist to view and interact with the model’s intermediate attribute predictions that support its final diagnosis. If the radiologist disagrees with any of these attributes, they can modify them, prompting the model to generate a revised target prediction based on that adjustment.

The target classification performance under incorrect attribute prediction is reported as Within-1-Accuracy (mean/std) for each attribute: 96.9\%/2.9\% (sub), 100.0\%/0.0\% (is), 52.6\%/11.6\% (cal), 93.0\%/4.3\% (sph), 93.8\%/4.2\% (mar), 93.5\%/4.3\% (lob), 89.7\%/3.5\% (spic), 98.1\%/1.0\% (tex). These values were obtained when measuring the target performance only for samples, where the attribute prediction of the respective attribute was incorrect. 
Higher values (e.g., is, tex) may be partially attributed to the limited number of failure cases per attribute, as well as a weak coupling between these attributes and the final target prediction. In contrast, stronger drops for calcification and spiculation indicate a more pronounced dependency between attribute correctness and target classification, reflecting stronger error propagation. The test-time intervention can mitigate this drop when the incorrectness of the attribute prediction is being noticed by the user.

In this experiment, the intervention is evaluated under an oracle condition, where the correct attribute score is assumed to be known. This provides an upper-bound analysis of the potential benefit of correcting erroneous attribute predictions.

For test-time intervention, we leverage prime prototype vectors, each corresponding to a specific attribute score. Whenever an attribute prediction is considered incorrect, its representation vector is replaced by the prime prototype vector associated with the correct attribute score. To obtain the prime prototypes, we identify, for each attribute score, the prototype that is selected most frequently as the nearest prototype among validation samples of that score. This ensures that the chosen prototype is representative of the majority of instances within its category.

To determine the correction order, we follow Koh et al.’s procedure on Concept Bottleneck Models \cite{koh2020concept}. Using the validation set, we measure the individual target prediction gain achieved by correcting each attribute in isolation. In our final experiment, we sequentially correct attributes in descending order of their measured gain, starting with the one that yields the largest improvement when corrected, followed by the next most impactful, and so on, until all incorrect attributes are addressed. 

\begin{figure}[!htbp]
    \centering
    \includegraphics[width=\linewidth]{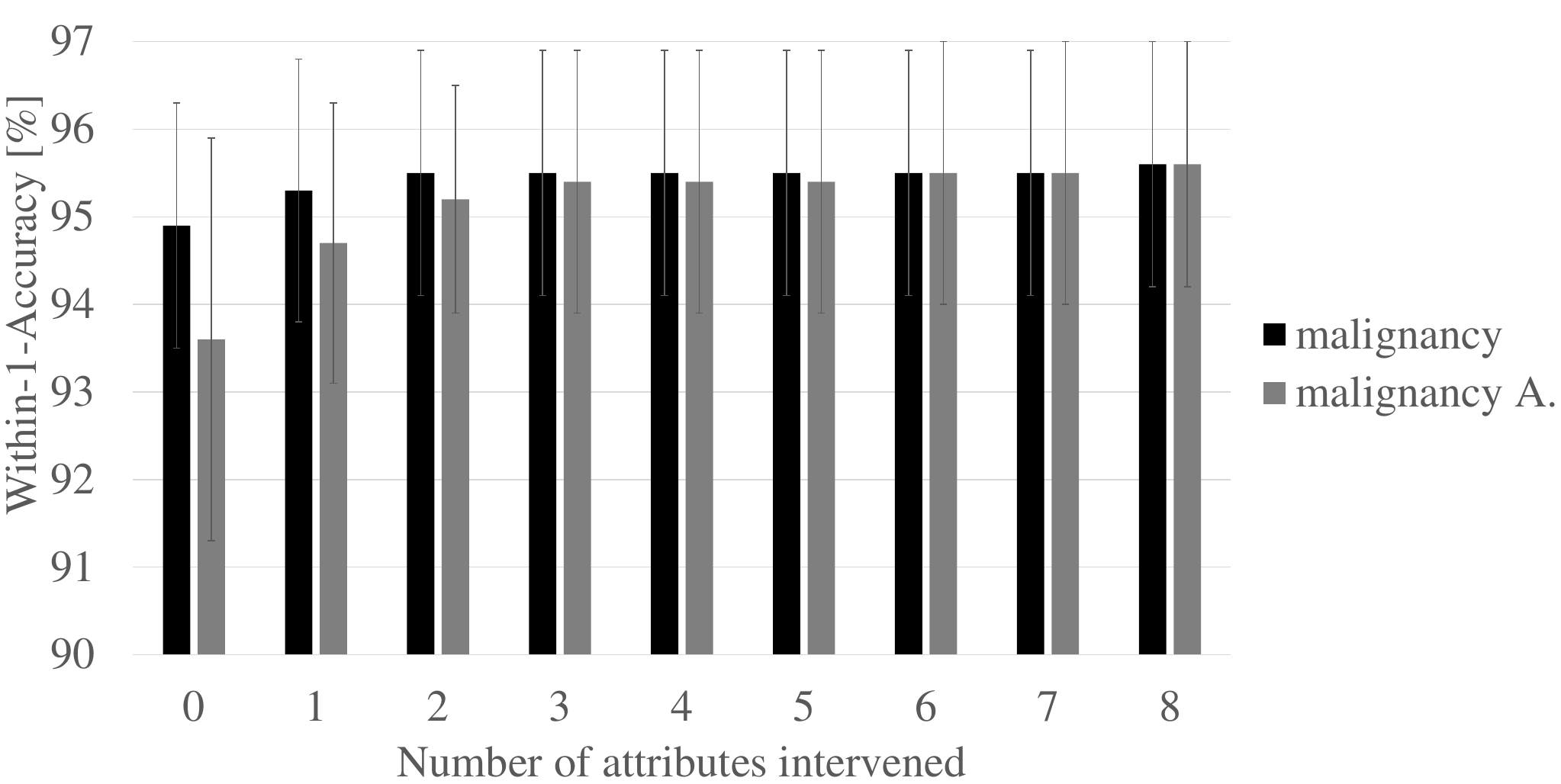}
    \caption{\textbf{Test-time intervention} Intervening in incorrect attribute predictions improves overall accuracy, with performance gains plateauing after three interventions. From this point onward, the ablated model using prototype vector-based inference for malignancy prediction (malignancy A.) achieves performance comparable to the original model. Error bars show the standard deviation of 5-fold experiments. 
    }
    \label{fig:tti}
\end{figure}

Figure \ref{fig:tti} shows the results of the test‑time intervention on the LIDC‑IDRI dataset. As successive attribute predictions are corrected, overall accuracy increases and then plateaus after three interventions. The maximum accuracy achieved is $95.6\%$ (SD=$1.4\%$). 

In ablated version A (see Section \ref{para:A.}), where every attribute vector is replaced by its nearest prototype, the accuracy converges to that of the original model, as shown by the 'malignancy A.' curve.

For four of the five folds, the most impactful attributes, and therefore the first to be intervened on, are spiculation, calcification, and lobulation. This pattern may reflect the relatively low baseline accuracy for spiculation, indicating that its correction could lead to the largest improvement in the target prediction.

The inference time of the proposed test-time intervention algorithm includes feature extraction by the backbone, processing through the attribute branches, replacement of an attribute vector with a prime prototype vector, and subsequent target classification. The overall runtime remains below 20\,ms per sample with a batch size of 1, demonstrating low inference-time latency under the reported experimental setting.

By exposing intermediate attribute predictions and enabling on‑the‑fly corrections, HierViT transforms passive inference into a collaborative process. Radiologists gain transparency into the model’s decision pathway and can directly test “what‑if” scenarios (for instance, “What if the lesion margin were classified as smooth instead of spiculated?”) to instantly assess the impact on malignancy prediction. This interactive loop not only enhances interpretability but also allows clinicians to compensate for known model weaknesses.

\subsection{Prototype analysis}
\begin{figure}[ht]
    \centering
    \begin{subfigure}{0.45\linewidth}
    \centering
        \includegraphics[height=0.5\linewidth]{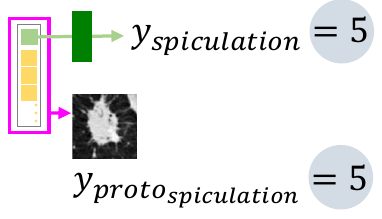}
        \caption{Prototype truthfulness}
    \end{subfigure}
    \hfill
    \begin{subfigure}{0.45\linewidth}
        \centering
        \includegraphics[height=0.6\linewidth]{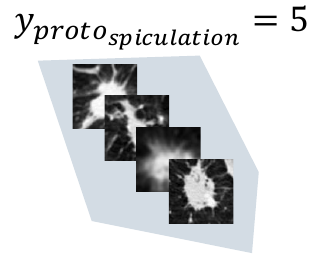}
        \caption{Prototype diversity}
    \end{subfigure}
    \begin{subfigure}{0.75\linewidth}
        \vspace{20pt}
        \includegraphics[width=\linewidth]{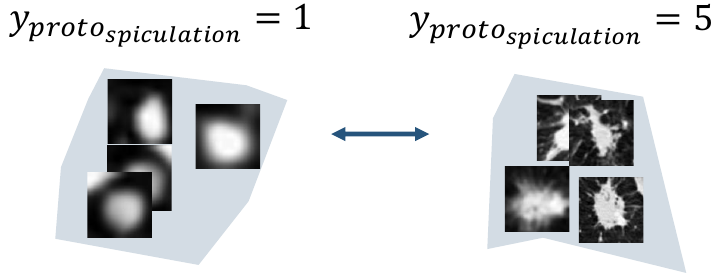}
        \caption{Counterfactual prototypes}
    \end{subfigure}
    \caption{\textbf{Analysis of learned prototypes} (a) agreement between prototypes, ground truth, and made predictions, (b) diversity of used prototypes, and (c) counterfactual prototypes.}
    \label{fig:protoanalysis}
\end{figure}

The goal of explainable AI methods is to achieve high prediction accuracy while providing understandable explanations. The accuracy of target and attribute predictions has been evaluated in previous experiments. This section focuses on analyzing the quality of HierViT's prototypical explanations. Inspired by Pathak \textit{et al.} \cite{pathak2024Prototype}, we assess the prototypes generated by HierViT in terms of truthfulness, diversity, and counterfactuality, as shown in Fig. \ref{fig:protoanalysis}.

\paragraph{Prototype truthfulness}
The model aims to generate visual prototypes that represent the recognized features. In HierViT's architecture, this means the prototypes reflect the attribute classification predictions. Specifically, the attributes of the selected prototype sample for each latent vector should align with the predicted attributes from these vectors. 

In the case of the LIDC-IDRI dataset, the following levels of agreement were determined for the attributes: $99.0\%$ (sub), $100.0\%$ (is), $99.2\%$ (cal), $97.0\%$ (sph), $97.7\%$ (mar), $99.9\%$ (lob), $95.5\%$ (spic), $96.1\%$ (tex).

For the derm7pt dataset, the following levels of agreement were determined for the attributes: $96.0\%$ (pn), $99.2\%$ (bmv), $98.5\%$ (vs), $96.2\%$ (pig), $98.5\%$ (str), $96.2\%$ (dag), $69.4\%$ (rs).

The agreement between the prototype values and the attribute predictions is predominantly high across all attributes, with the exception of the regression structures attribute, where a lower agreement is observed. 

This suggests that prototype learning for this attribute is less stable, resulting in reduced faithfulness of the corresponding prototype-based explanations compared to the remaining attributes.

\paragraph{Prototype diversity}
The diversity of generated prototypes indicates whether they provide local or global explanations. Local explainability highlights sample-specific features, allowing different prototypes for inference samples of the same class. In contrast, global explainability offers class-specific insights by generating cluster prototypes that effectively differentiate between classes.
In the following experiment, we analyzed prototype utilization during inference on the test splits. A prototype was considered \textit{used} if it was selected as the nearest prototype for at least one test sample. For each attribute class, we counted the number of distinct prototypes that were used and averaged this count across all attribute classes. A set of sixteen prototypes per attribute class was provided for both experiments on the LIDC-IDRI and on the derm7pt dataset. This choice was intended to provide sufficient capacity to capture intra-class variation rather than to match the expected number of utilized prototypes.
On average, 3.8 of the 16 available prototypes per attribute class were used in the LIDC-IDRI experiment, compared to 7.6 of 16 in the derm7pt experiment.

The latter result indicates a higher degree of diversity and differentiation among the features of the same attribute class.
For the LIDC-IDRI experiment, this indicates that only a subset of the available prototypes is required to represent the test samples, suggesting redundancy among prototypes or concentration of samples in specific regions of the attribute space. 
The loss term $L_{proto}$ in Equation \ref{eq:L_proto} steers the adaptation of the prototype vectors and influences the latent space of the prototypes. 
In contrast to \cite{gallee2023interpretable}, which applies the loss term only to the nearest prototypes and reports a higher attribute-prototype diversity of 7.8 on LIDC-IDRI, we employ a holistic adaptation of all prototypes belonging to the correct class, motivated by improved prototype-inference performance.
An additional hyperparameter that may influence the number of effectively used prototypes is the number of available prototypes per attribute class $P$ \cite{gallee2026protocaps}. 
Adapting the model to the current setting with respect to these design choices enables a different balance between prototype diversity and performance.

\paragraph{Counterfactual Prototypes}
Counterfactual prototypes are those furthest from the inference sample in the model's latent space. 
For instance, in the LIDC-IDRI dataset, prototypes representing 'linear' and 'round' should be distant for the attribute \textit{sphericity}. Consequently, employing counterfactual prototypes should result in changed predictions. This effect is measured using the inverse accuracy (1-ACC) for attribute classification with counterfactual prototypes.

For the LIDC-IDRI dataset, the following inverse accuracies were measured on average over the 5-fold experiments: $97.0\%$ (sub), $0.2\%$ (is), $99.7\%$ (cal), $97.2\%$ (sph), $90.9\%$ (mar), $34.5\%$ (lob), $56.7\%$ (spic), $77.0\%$ (tex). The \textit{internal structure} (is) attribute exhibits a very low level of inverse accuracy, which can be attributed to a significant degree of data imbalance with regard to one class (with $98.9\%$ of data points occurring for class 1).

For the derm7pt dataset, the following inverse accuracies were measured: $91.4\%$ (pn), $86.8\%$ (bmv), $93.2\%$ (vs), $92.4\%$ (pig), $91.9\%$ (str), $86.3\%$ (dag), $36.0\%$ (rs).

Lower inverse accuracies, observed for attributes such as lobulation and spiculation on LIDC-IDRI and regression structures on derm7pt, indicate a less distinct separation of opposing attribute classes in the latent space. For most attributes, however, the experiments yield high inverse accuracies, suggesting that the model effectively separates prototypes associated with different attribute classes.

\subsection{Sparse data study}
The HierViT learning algorithm uses human-defined attributes as domain knowledge for reasoning. We initially tested it on densely annotated datasets with both target and attribute ground truth labels. However, generating such labels in the medical field is time-consuming and costly due to the need for expert knowledge.
To evaluate HierViT in scenarios with fewer annotations, we conducted experiments using the LIDC-IDRI dataset with sparse attribute labels. We tested HierViT with and without prototype inference, selecting random subsets of the data: $1\%$, $5\%$, $10\%$, and $50\%$ of samples with full annotations, while the remaining samples lacked attribute labels and only retained target labels.
\begin{figure}[!htbp]
    \centering
    \includegraphics[width=\linewidth]{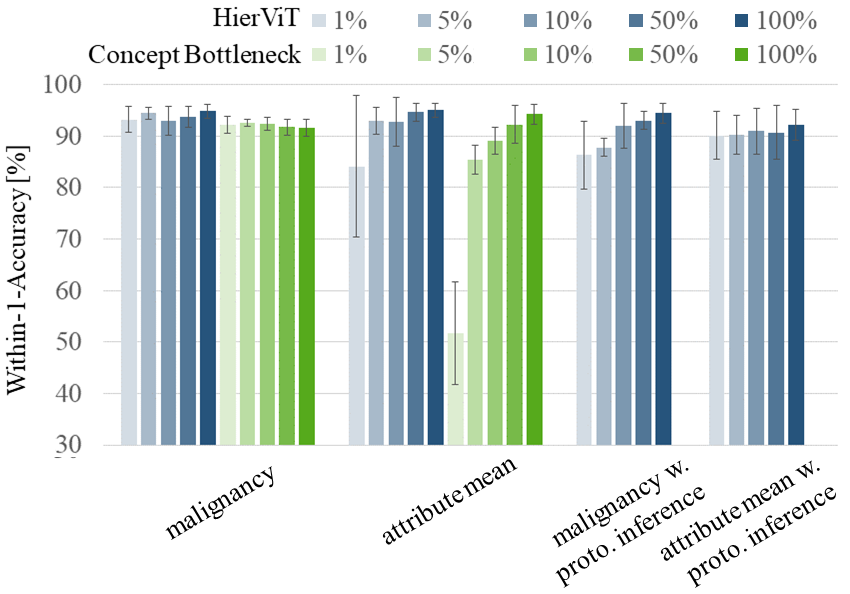}
    \caption{\textbf{Sparse data study} Performance results with reduced availability of attribute annotations using the LIDC-IDRI dataset. Error bars show the standard deviation of 5-fold experiments. 
    }
    \label{fig:sparsedatastudy}
\end{figure}

As demonstrated in Fig. \ref{fig:sparsedatastudy}, our results indicate that comparable performance can be achieved with reduced annotations. For example, using only $10\%$ of the attribute labels (1,907 training samples), we obtained malignancy prediction performance of 92.9/2.8, compared to 94.8/1.4 with $100\%$ of the labels, and attribute means performance of 92.7/4.7 versus 95.0/1.4 in the proposed method.

The Concept Bottleneck model \cite{koh2020concept} experiences a larger performance drop in attribute prediction as fewer annotations are available. 
This reduced robustness may be partly attributed to the absence of an explicit prototype-learning mechanism, as also indicated by Proto-Caps \cite{gallee2023interpretable}, which maintains improved stability under low-annotation regimes and thereby supports the effectiveness of attribute-level prototypes in limited-data settings.

These findings suggest that while HierViT benefits from domain knowledge, it does not depend on densely annotated datasets, enhancing its practicality and applicability to other datasets with limited annotations. This robustness may facilitate deployment in other medical imaging domains, where expert-defined attribute annotations are scarce due to the substantial time and expertise required for their collection, while still enabling attribute-based and prototype-based explanations.

\paragraph{Synthesis of attribute-annotated data}
To compensate for sparse ground-truth attributes, we synthesized attribute-labeled data to enrich the training set for HierViT, as described in a separate work \cite{gallee2025minimum}. We developed a Diffusion Model with a U‑Net backbone and attribute‑conditioning via cross‑attention in the bottleneck, and it was trained using a semi‑supervised approach.
We augmented the real dataset with these synthetic examples and demonstrated that using only 20 fully annotated real cases yields a malignancy prediction accuracy of 92.7/1.6 and an attribute mean accuracy of 93.5/2.0. Although this performance is somewhat lower than that obtained on the densely annotated dataset, the annotation burden is dramatically reduced (20 samples versus 19,713 samples). 

\subsection{Hyperparameter-Analysis}
\label{subsec:hyperparameter_analysis}

\begin{figure}[!htbp]
    \centering
    \includegraphics[width=\linewidth]{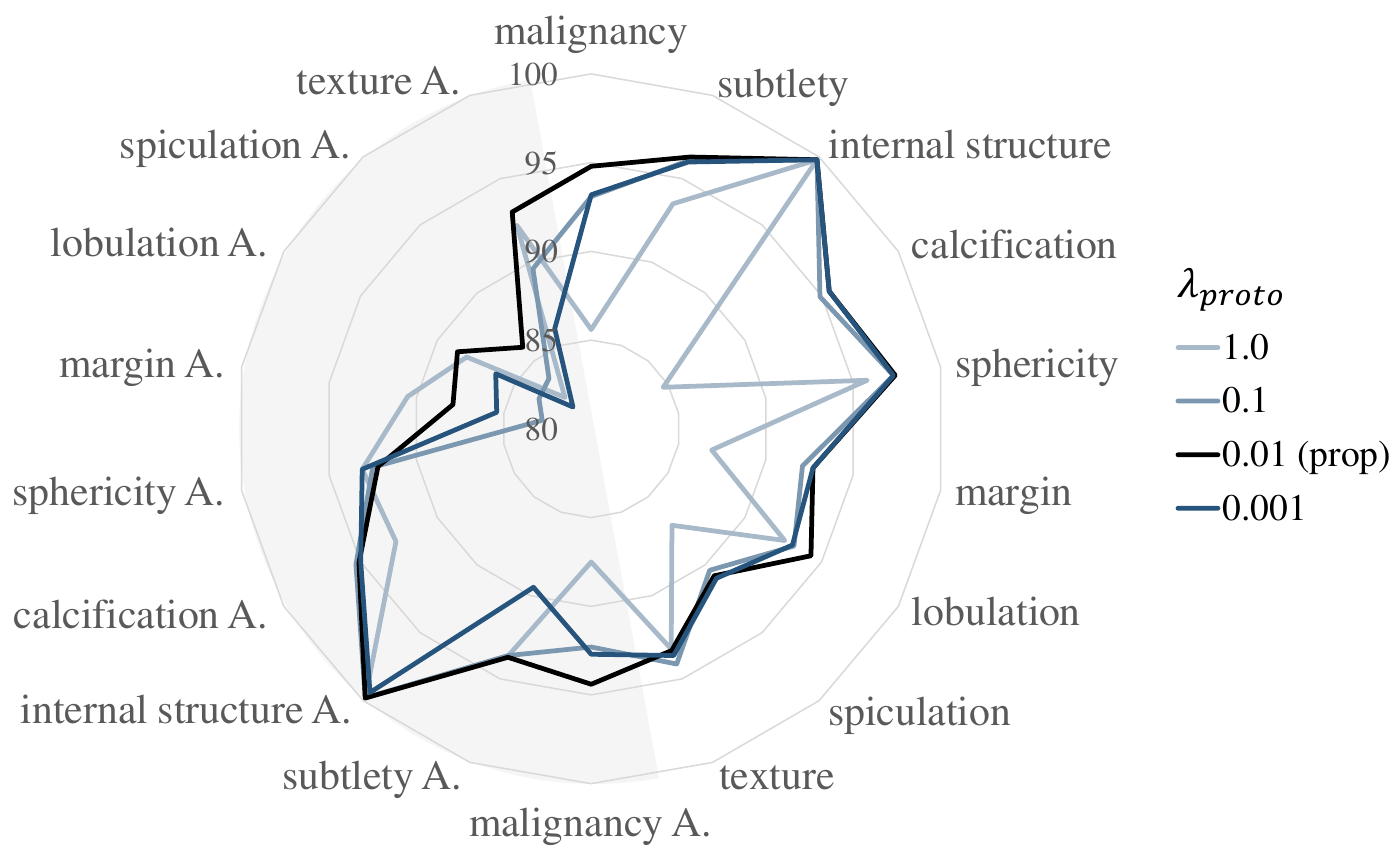}
    \caption{Evaluation of the parameter $\lambda_{proto}$ in the combined loss Equation \ref{eq:combined_loss_function} on the LIDC-IDRI dataset. Labels with the suffix 'A.' (gray background) indicate results where prototypes were used during inference (see Section \ref{para:A.} for details). 
    }
    \label{fig:lamda_proto}
\end{figure}
To assess the influence of the weighting factor $\lambda_{proto}$ in the combined loss function, we conducted a series of experiments varying its value. Specifically, we evaluated $\lambda_{proto} \in$ \{0.001, 0.01, 0.1, 1.0\}. This parameter controls the contribution of the prototype-related loss component relative to the primary task loss, see Section \ref{subsec:trainingalgoLOSS}.

Our results (see Figure~\ref{fig:lamda_proto}) show that setting $\lambda_{\text{proto}} = 0.01$ yields the highest values for the majority of targets. This value allows the model to effectively leverage prototype information without compromising the primary task objective. In the figure, the right part displays the performance of the proposed HierViT algorithm, while the left part presents ablated results, indicated by labels with the suffix 'A.', which correspond to models that use prototype latent vectors during inference. For a detailed explanation, see Section~\ref{para:A.}.

Based on this analysis, we selected $\lambda_{proto} = 0.01$ for all subsequent experiments. This choice maintains the model’s focus on the main task while still benefiting from prototype-guided regularization.

\section{Discussion}
HierViT advances Vision Transformers in explainable AI by leveraging domain knowledge and a hierarchical architecture for human-like reasoning. Experiments on the LIDC-IDRI and derm7pt benchmark datasets demonstrate competitive performance alongside enhanced explainability. The model infers target predictions from recognized visual features using ratings (e.g., “The shape is round and the texture is solid”), prototypical sample images (e.g., “The sphericity is similar to this sample”), and attention heatmaps (e.g., “This area is crucial for recognizing sphericity”). These human-defined attributes are understandable and learned in a supervised manner, providing interpretable information for target outputs as they feed into the target prediction branch.

The proposed model architecture allows for substantial domain knowledge integration. However, ablation studies show that HierViT maintains comparable performance even when certain components and annotations are omitted. For instance, it performs robustly with a $90\%$ reduction in attribute annotations, indicating its applicability to other datasets with limited supervision. This suggests that the framework can also be adapted to more realistic clinical scenarios, where attribute annotations may be incomplete. 

Furthermore, on the derm7pt dataset, which lacks region-of-interest maps, as well as in the corresponding ablation study on LIDC-IDRI without segmentation supervision, the model maintains competitive performance, indicating robustness to missing spatial annotations. On the other hand, improving the optimization of the decoder branch for more accurate segmentation could be an interesting direction for future work, particularly to investigate its impact on the model’s localization ability and its relationship to the attribute-specific attention maps. Additionally, while self-attention maps provide an explainability mechanism, quantitatively evaluating their faithfulness against attribute-level ground-truth annotations remains an important direction for future work.

Performance variations across attribute classes are observed under class imbalance, particularly in the derm7pt dataset, where minority attributes, such as the \textit{irregular} class of vascular structures (vs), show reduced performance.
As the target classifier operates on the latent representations of the attribute branches rather than their discrete predictions, errors in individual attribute classifications do not directly translate into target prediction errors. However, inaccurate attribute representations may still influence the target prediction, which can be desirable when attributes are clinically related to the target and contribute to its interpretable reasoning.
The lower agreement between predicted attributes and retrieved prototypes for the \textit{regression structures} (rs) attribute in the derm7pt experiment further highlights the impact of class imbalance on prototype-based explanations. Class imbalance represents a key challenge in real-world clinical datasets and needs to be addressed explicitly.
Future work may mitigate this limitation through oversampling techniques or attribute-aware synthetic data generation methods \cite{gallee2025minimum}, thereby increasing the availability of underrepresented attribute classes.

Test-time intervention on incorrect attributes can improve target prediction by enabling explicit correction of intermediate attribute prototypes. This allows users to interactively explore model behavior and gain insight into its decision process. However, a systematic evaluation of intervention faithfulness (i.e., whether changes in an attribute consistently induce logically coherent changes in the final prediction) remains an important direction for future work and can be rigorously studied using controlled synthetic datasets such as \cite{gallee2026funnynodules}.

Intuitive reasoning can enhance the perceived interpretability of the model by using predefined attributes aligned with human language, which may affect radiologists' diagnostic accuracy positively or negatively. A user-centered study on reasoning by attribute prototypes \cite{gallee2024evaluating} found that such explanations can increase radiologists' diagnostic confidence. 
However, the proposed explainability framework should be regarded as a technical approach for generating interpretable model outputs rather than a clinically validated tool. Its clinical usefulness and impact on end users remain to be established before real-world deployment.
In particular, systematic evaluation is required to assess the effect of attribute scores, prototypes, and attention-based explanations in AI-assisted diagnosis in terms of diagnostic accuracy, decision time, and trust calibration of clinicians.

The prototype analysis demonstrates that the predicted prototype samples accurately represent the identified attributes and effectively distinguish from counterfactuals. Additionally, using attribute prototypes for target prediction inference yields high accuracy, indicating an effective prototype learning strategy. However, as the current prototype objective does not explicitly account for prototype diversity, some prototypes may capture overlapping aspects of intra-class variation.

\textit{Limitations:}
While specific predefined attributes are crucial for attribute-based models in medical applications, there is a scarcity of medical datasets with such discrete annotations, highlighting the need for further research. There is potential to transfer attribute knowledge to radiological diagnosis tasks with similar visual criteria. The lack of datasets with annotated attributes in medicine also poses a challenge for model evaluation, limiting the assessment of the generalizability of attribute-based methods, including our approach, across different datasets, institutions, and imaging protocols. Moreover, as the predefined attributes are based on expert-defined criteria, their interpretation may be subject to inter-observer variability.

Another interesting area of research is the compensation for limited or absent attribute annotations. The transformer architecture could be extended by integrating a text processing branch to fuse radiology reports with image data, thereby incorporating richer domain knowledge in addition to, or potentially as an alternative to, attribute annotations. 
More generally, the integration of additional clinical metadata, such as patient information available in derm7pt but not used in this study, represents a promising direction for future work to incorporate complementary clinical information and better reflect the multi-modal reasoning used in clinical practice.
Such information could be embedded as feature tokens and integrated into the Transformer sequence alongside the visual tokens, enabling the model to jointly process visual and clinical information.

Another important direction for future clinical application is the integration of this approach with a preceding detection or segmentation module to automatically identify relevant regions of interest in full medical images, as the lung nodules and skin lesions in this study are provided as pre-cropped regions.


\section{Conclusion}
This work presents an image classifier that incorporates multiple interpretable modalities. It utilises a Vision Transformer as a high-performing backbone and a hierarchical classifier. In the initial step, diagnostic criteria specified by human experts are learned. Based on these recognized attributes, the target classification is made. 
This step-by-step structure enables the establishment of a semantic relationship between the attributes and the target result. 
The explainability of the model is provided by the detailed description of the recognized attributes, including ratings, visual prototypes, and attention heatmaps. The model demonstrates which known example image is closest to an attribute and which area it has focused on in particular. This approach provides structured intermediate representations that can support more intuitive interpretation of the model’s predictions.

\appendix
\section{Appendix}
\label{app1}

\begin{table}[!htbp]
  \caption{\textbf{derm7pt: detailed target performance comparison} The target (diag) performance of the experiment on the derm7pt dataset is reported in sensitivity (sen), specificity (spe), precision (pre), and Area Under the Curve (auc). In the HierViT's category using dermoscopic images only, the highest values are in bold.}
  \label{dermtarget}
  \centering
  \begin{tabular}{lcccccc}
    \toprule
    &&\multicolumn{5}{c}{diag}\\
    \cmidrule(lr{0.5em}){3-7}
    &&nev&sk&misc&bcc&mel\\
    \midrule
    \multirow{2}{*}{\makecell[l]{\textbf{using dermoscopic}\\\textbf{images only}}}&&&&&&\\
    &&&&&&\\
    \multirow{2}{*}{\makecell[l]{AMFAM\\ derm. only \cite{wang2022adversarial}}}
    &sen&73.0&0.0&34.8&14.3&\textbf{71.7}\\
    &spe&\textbf{88.5}&95.2&91.4&96.1&86.5\\
    &pre&\textbf{94.1}&0.0&20.0&6.3&58.4\\
    &auc&88.4&73.6&89.1&90.5&85.2\\
    \midrule
    \multirow{2}{*}{\makecell[l]{HcCCN\\ derm. only \cite{bi2020multi}}}
    &sen&-&-&-&-&-\\
    &spe&-&-&-&-&-\\
    &pre&-&-&-&-&-\\
    &auc&82.7&79.2&87.1&88.3&80.3\\
    \midrule
    Concept Bottleneck\cite{koh2020concept}
    &sen&87.7&5.3&60.0&25.0&56.4\\
    &spe&68.2&99.2&93.8&\textbf{97.9}&\textbf{90.5}\\
    &pre&77.4&25.0&52.2&33.3&67.1\\
    &auc&85.0&80.1&88.1&89.1&83.6\\
    \midrule
    HierViT (proposed)
    &sen&\textbf{91.3}&\textbf{15.8}&\textbf{65.0}&\textbf{43.8}&65.3\\
    &spe&67.5&\textbf{99.7}&\textbf{97.2}&97.4&89.7\\
    &pre&80.3&\textbf{75.0}&\textbf{76.5}&\textbf{46.7}&\textbf{71.0}\\
    &auc&\textbf{90.6}&\textbf{82.4}&\textbf{91.4}&\textbf{97.8}&\textbf{89.9}\\
    \cmidrule{1-1}
    \multirow{3}{*}{\makecell[l]{\textcolor{gray}{\textbf{using dermoscopic}}\\ \textcolor{gray}{\textbf{+ clinical images,}}\\\textcolor{gray}{\textbf{ and metadata*}}}}&&&&&&\\
    &&&&&&\\
    &&&&&&\\
    \textcolor{gray}{AMFAM \cite{wang2022adversarial}}
    &\textcolor{gray}{sen}&\textcolor{gray}{84.1}&\textcolor{gray}{40.0}&\textcolor{gray}{68.0}&\textcolor{gray}{40.0}&\textcolor{gray}{65.8}\\
    &\textcolor{gray}{spe}&\textcolor{gray}{85.8}&\textcolor{gray}{95.6}&\textcolor{gray}{93.8}&\textcolor{gray}{97.4}&\textcolor{gray}{91.4}\\
    &\textcolor{gray}{pre}&\textcolor{gray}{89.5}&\textcolor{gray}{10.0}&\textcolor{gray}{42.5}&\textcolor{gray}{37.5}&\textcolor{gray}{76.2}\\
    &\textcolor{gray}{auc}&\textcolor{gray}{89.7}&\textcolor{gray}{81.7}&\textcolor{gray}{90.6}&\textcolor{gray}{94.1}&\textcolor{gray}{89.1}\\
    \midrule 
    \textcolor{gray}{HcCCN \cite{bi2020multi}}
    &\textcolor{gray}{sen}&\textcolor{gray}{-}&\textcolor{gray}{-}&\textcolor{gray}{-}&\textcolor{gray}{-}&\textcolor{gray}{58.4}\\
    &\textcolor{gray}{spe}&\textcolor{gray}{-}&\textcolor{gray}{-}&\textcolor{gray}{-}&\textcolor{gray}{-}&\textcolor{gray}{88.1}\\
    &\textcolor{gray}{pre}&\textcolor{gray}{-}&\textcolor{gray}{-}&\textcolor{gray}{-}&\textcolor{gray}{-}&\textcolor{gray}{62.8}\\
    &\textcolor{gray}{auc}&\textcolor{gray}{87.7}&\textcolor{gray}{80.4}&\textcolor{gray}{88.3}&\textcolor{gray}{94.4}&\textcolor{gray}{85.6}\\
    \midrule
    \multirow{3}{*}{\makecell[l]{\textcolor{gray}{Inception-}\\\textcolor{gray}{combine*}\\ \textcolor{gray}{\cite{kawahara2019seven}}}}
    &\textcolor{gray}{sen}&\textcolor{gray}{91.3}&\textcolor{gray}{15.8}&\textcolor{gray}{42.5}&\textcolor{gray}{25.0}&\textcolor{gray}{55.4}\\
    &\textcolor{gray}{spe}&\textcolor{gray}{62.5}&\textcolor{gray}{99.7}&\textcolor{gray}{97.2}&\textcolor{gray}{98.9}&\textcolor{gray}{88.4}\\
    &\textcolor{gray}{pre}&\textcolor{gray}{75.2}&\textcolor{gray}{75.0}&\textcolor{gray}{63.0}&\textcolor{gray}{50.5}&\textcolor{gray}{62.2}\\
    &\textcolor{gray}{auc}&\textcolor{gray}{88.1}&\textcolor{gray}{90.4}&\textcolor{gray}{86.8}&\textcolor{gray}{89.2}&\textcolor{gray}{84.2}\\
    \midrule   
    \textcolor{gray}{FusionM4Net* \cite{tang2022Fusion}}
    &\textcolor{gray}{sen}&\textcolor{gray}{95.0}&\textcolor{gray}{10.5}&\textcolor{gray}{52.5}&\textcolor{gray}{43.8}&\textcolor{gray}{71.3}\\
    &\textcolor{gray}{spe}&\textcolor{gray}{73.3}&\textcolor{gray}{99.7}&\textcolor{gray}{98.9}&\textcolor{gray}{97.9}&\textcolor{gray}{91.5}\\
    &\textcolor{gray}{pre}&\textcolor{gray}{81.6}&\textcolor{gray}{66.7}&\textcolor{gray}{84.0}&\textcolor{gray}{46.7}&\textcolor{gray}{74.2}\\
    &\textcolor{gray}{auc}&\textcolor{gray}{95.4}&\textcolor{gray}{91.5}&\textcolor{gray}{95.0}&\textcolor{gray}{95.4}&\textcolor{gray}{92.6}\\
    \bottomrule
  \end{tabular}
\end{table}

\begin{landscape}
\begin{table}[!htbp]
  \caption{\textbf{derm7pt: detailed attribute performance comparison} The attribute performance of the experiment on the derm7pt dataset is reported in sensitivity (sen), specificity (spe), precision (pre), and Area Under the Curve (auc). In the HierViT's category using dermoscopic images only, the highest values are in bold.}
  \label{dermattr}
  \centering
  \footnotesize
  \begin{tabular}{lcccccccccccccccccccc}
    \toprule
    &&\multicolumn{3}{c}{pn}&\multicolumn{2}{c}{bmv}&\multicolumn{3}{c}{vs}&\multicolumn{3}{c}{pig}&\multicolumn{3}{c}{str}&\multicolumn{3}{c}{dag}&\multicolumn{2}{c}{rs}\\
    \cmidrule(lr{0.5em}){3-5}
    \cmidrule(lr{0.5em}){6-7}
    \cmidrule(lr{0.5em}){8-10}
    \cmidrule(lr{0.5em}){11-13}
    \cmidrule(lr{0.5em}){14-16}
    \cmidrule(lr{0.5em}){17-19}
    \cmidrule(lr{0.5em}){20-21}
    &&abs&typ&aty&abs&pre&abs&reg&irg&abs&reg&irg&abs&reg&irg&abs&reg&irg&abs&pre\\
    \midrule
    \multirow{2}{*}{\makecell[l]{\textbf{using dermoscopic}\\\textbf{images only}}}&&&&&&\\
    &&&&&&\\
    \multirow{2}{*}{\makecell[l]{AMFAM\\ derm. only\cite{wang2022adversarial}}}
    &sen&73.3&74.2&\textbf{46.4}&87.2&\textbf{85.3}&83.1&48.3&0.0&65.1&\textbf{72.2}&\textbf{73.6}&75.5&1.0&52.4&\textbf{61.7}&52.8&64.8&83.1&\textbf{62.1}\\
    &spe&83.8&80.1&85.3&\textbf{85.3}&87.3&\textbf{69.0}&89.6&92.4&\textbf{83.1}&90.7&75.1&\textbf{68.8}&90.9&83.9&81.2&\textbf{83.0}&74.3&\textbf{62.1}&83.1\\
    &pre&\textbf{75.6}&63.0&54.8&\textbf{98.4}&38.7&\textbf{97.1}&26.9&0.0&\textbf{94.6}&27.1&31.5&\textbf{88.7}&20.5&46.8&37.0&\textbf{63.6}&\textbf{70.6}&\textbf{88.6}&50.9\\
    &auc&86.9&81.4&77.1&\textbf{87.4}&\textbf{87.4}&84.2&85.3&\textbf{71.1}&\textbf{79.2}&\textbf{82.9}&\textbf{79.6}&79.7&\textbf{86.9}&77.1&\textbf{76.7}&75.8&77.2&\textbf{80.0}&\textbf{80.0}\\
    \midrule
    \multirow{2}{*}{\makecell[l]{HcCCN\\ derm. only \cite{bi2020multi}}}
    &sen&-&-&-&-&-&-&-&-&-&-&-&-&-&-&-&-&-&-&-\\
    &spe&-&-&-&-&-&-&-&-&-&-&-&-&-&-&-&-&-&-&-\\
    &pre&-&-&-&-&-&-&-&-&-&-&-&-&-&-&-&-&-&-&-\\
    &auc&-&83.5&74.5&-&82.6&-&83.1&66.7&-&73.7&79.2&-&81.2&71.6&-&72.0&\textbf{78.0}&-&75.0\\
    \midrule
    Concept Bottleneck \cite{koh2020concept}
    &sen&73.1&\textbf{83.6}&31.2&93.8&52.0&\textbf{92.0}&51.9&\textbf{6.7}&\textbf{86.5}&25.0&50.1&\textbf{87.5}&\textbf{50.0}&41.5&38.0&\textbf{66.9}&64.4&88.6&45.3\\
    &spe&\textbf{84.5}&71.1&\textbf{93.0}&52.0&93.8&51.2&\textbf{90.7}&\textbf{98.4}&59.3&94.8&\textbf{85.6}&61.6&94.3&\textbf{88.0}&\textbf{89.5}&71.5&\textbf{75.2}&45.3&88.6\\
    &pre&75.5&62.9&\textbf{58.0}&89.3&66.1&87.8&\textbf{45.8}&\textbf{25.0}&73.4&40.0&\textbf{61.8}&80.9&52.4&52.0&\textbf{55.1}&50.0&67.9&81.5&59.3\\
    &auc&\textbf{89.4}&\textbf{84.7}&76.5&68.7&31.3&36.2&35.5&60.2&41.3&49.5&69.4&39.4&36.4&72.2&66.1&70.8&71.7&44.9&55.1\\ 
     
    \midrule
    \multirow{2}{*}{\makecell[l]{HierViT\\(proposed)}}
    &sen&\textbf{80.8}&64.4&40.9&\textbf{94.7}&57.3&91.4&\textbf{57.7}&3.3&79.4&33.3&58.9&85.6&43.2&\textbf{58.5}&32.0&\textbf{66.9}&\textbf{69.5}&\textbf{91.3}&48.1\\
    &spe&69.5&\textbf{80.8}&84.6&57.3&\textbf{94.7}&47.0&88.3&\textbf{98.4}&57.8&\textbf{96.2}&78.1&60.2&\textbf{96.5}&85.1&88.2&68.6&63.8&48.1&\textbf{91.3}\\
    &pre&68.5&\textbf{70.7}&48.7&90.4&\textbf{71.7}&89.1&44.1&16.7&73.1&\textbf{61.5}&57.5&81.8&\textbf{65.5}&\textbf{56.7}&54.2&52.7&66.1&82.8&\textbf{67.1}\\
    &auc&86.7&81.9&\textbf{79.4}&86.0&86.0&\textbf{88.6}&\textbf{87.7}&54.3&77.1&76.2&77.2&\textbf{81.8}&82.1&\textbf{79.2}&68.3&\textbf{76.4}&76.3&\textbf{80.0}&\textbf{80.0}\\
    \cmidrule{1-1}
    \multirow{3}{*}{\makecell[l]{\textcolor{gray}{\textbf{using dermoscopic}}\\ \textcolor{gray}{\textbf{+ clinical images,}}\\\textcolor{gray}{\textbf{ and metadata*}}}}&&&&&&\\
    &&&&&&\\
    &&&&&&\\
    \multirow{2}{*}{\makecell[l]{\textcolor{gray}{AMFAM}\\\textcolor{gray}{\cite{wang2022adversarial}}}}
    &\textcolor{gray}{sen}&\textcolor{gray}{77.2}&\textcolor{gray}{70.3}&\textcolor{gray}{58.5}&\textcolor{gray}{90.3}&\textcolor{gray}{75.0}&\textcolor{gray}{83.8}&\textcolor{gray}{75.0}&\textcolor{gray}{0.0}&\textcolor{gray}{72.7}&\textcolor{gray}{63.2}&\textcolor{gray}{67.9}&\textcolor{gray}{80.6}&\textcolor{gray}{70.4}&\textcolor{gray}{57.3}&
    \textcolor{gray}{68.4}&\textcolor{gray}{56.3}&\textcolor{gray}{66.7}&\textcolor{gray}{82.6}&\textcolor{gray}{72.1}\\
    &\textcolor{gray}{spe}&\textcolor{gray}{85.7}&\textcolor{gray}{84.6}&\textcolor{gray}{85.6}&\textcolor{gray}{75.0}&\textcolor{gray}{90.3}&\textcolor{gray}{91.7}&\textcolor{gray}{90.8}&\textcolor{gray}{92.4}&\textcolor{gray}{76.3}&\textcolor{gray}{90.4}&\textcolor{gray}{83.0}&\textcolor{gray}{72.4}&\textcolor{gray}{93.2}&\textcolor{gray}{85.9}&
    \textcolor{gray}{82.0}&\textcolor{gray}{81.5}&\textcolor{gray}{82.4}&\textcolor{gray}{72.1}&\textcolor{gray}{82.6}\\
    &\textcolor{gray}{pre}&\textcolor{gray}{78.2}&\textcolor{gray}{74.7}&\textcolor{gray}{51.6}&\textcolor{gray}{95.6}&\textcolor{gray}{56.0}&\textcolor{gray}{99.4}&\textcolor{gray}{34.6}&\textcolor{gray}{0.0}&\textcolor{gray}{86.1}&\textcolor{gray}{25.0}&\textcolor{gray}{61.3}&\textcolor{gray}{87.5}&\textcolor{gray}{43.2}&\textcolor{gray}{54.3}&
    \textcolor{gray}{39.0}&\textcolor{gray}{56.8}&\textcolor{gray}{82.5}&\textcolor{gray}{93.4}&\textcolor{gray}{46.2}\\
    &\textcolor{gray}{auc}&\textcolor{gray}{89.2}&\textcolor{gray}{84.5}&\textcolor{gray}{82.0}&\textcolor{gray}{91.1}&\textcolor{gray}{91.1}&\textcolor{gray}{89.6}&\textcolor{gray}{88.8}&\textcolor{gray}{80.9}&\textcolor{gray}{80.7}&\textcolor{gray}{85.1}&\textcolor{gray}{83.4}&\textcolor{gray}{83.0}&\textcolor{gray}{89.5}&\textcolor{gray}{80.7}&
    \textcolor{gray}{77.1}&\textcolor{gray}{77.7}&\textcolor{gray}{81.9}&\textcolor{gray}{86.7}&\textcolor{gray}{86.7}\\
    \midrule
    \multirow{2}{*}{\makecell[l]{\textcolor{gray}{HcCCN}\\\textcolor{gray}{\cite{bi2020multi}}}}
    &\textcolor{gray}{sen}&\textcolor{gray}{-}&\textcolor{gray}{-}&\textcolor{gray}{40.9}&\textcolor{gray}{-}&\textcolor{gray}{92.2}&\textcolor{gray}{-}&\textcolor{gray}{-}&\textcolor{gray}{20.0}&\textcolor{gray}{-}&\textcolor{gray}{-}&\textcolor{gray}{55.7}&\textcolor{gray}{-}&\textcolor{gray}{-}&\textcolor{gray}{35.1}&\textcolor{gray}{-}&\textcolor{gray}{-}&\textcolor{gray}{80.2}&\textcolor{gray}{-}&\textcolor{gray}{95.2}\\
    &\textcolor{gray}{spe}&\textcolor{gray}{-}&\textcolor{gray}{-}&\textcolor{gray}{92.4}&\textcolor{gray}{-}&\textcolor{gray}{65.3}&\textcolor{gray}{-}&\textcolor{gray}{-}&\textcolor{gray}{98.4}&\textcolor{gray}{-}&\textcolor{gray}{-}&\textcolor{gray}{86.3}&\textcolor{gray}{-}&\textcolor{gray}{-}&\textcolor{gray}{90.0}&\textcolor{gray}{-}&\textcolor{gray}{-}&\textcolor{gray}{71.6}&\textcolor{gray}{-}&\textcolor{gray}{41.5}\\
    &\textcolor{gray}{pre}&\textcolor{gray}{-}&\textcolor{gray}{-}&\textcolor{gray}{62.3}&\textcolor{gray}{-}&\textcolor{gray}{91.9}&\textcolor{gray}{-}&\textcolor{gray}{-}&\textcolor{gray}{50.0}&\textcolor{gray}{-}&\textcolor{gray}{-}&\textcolor{gray}{65.1}&\textcolor{gray}{-}&\textcolor{gray}{-}&\textcolor{gray}{52.4}&\textcolor{gray}{-}&\textcolor{gray}{-}&\textcolor{gray}{69.6}&\textcolor{gray}{-}&\textcolor{gray}{81.6}\\
    &\textcolor{gray}{auc}&\textcolor{gray}{-}&\textcolor{gray}{85.9}&\textcolor{gray}{78.3}&\textcolor{gray}{-}&\textcolor{gray}{89.8}&\textcolor{gray}{-}&\textcolor{gray}{87.0}&\textcolor{gray}{82.7}&\textcolor{gray}{-}&\textcolor{gray}{83.6}&\textcolor{gray}{81.3}&\textcolor{gray}{-}&\textcolor{gray}{87.8}&\textcolor{gray}{77.6}&\textcolor{gray}{-}&\textcolor{gray}{77.7}&\textcolor{gray}{82.6}&\textcolor{gray}{-}&\textcolor{gray}{81.9}\\
    \midrule
    \multirow{3}{*}{\makecell[l]{\textcolor{gray}{Inception-}\\\textcolor{gray}{combine*}\\ \textcolor{gray}{\cite{kawahara2019seven}}}}
    &\textcolor{gray}{sen}&\textcolor{gray}{77.6}&\textcolor{gray}{78.1}&\textcolor{gray}{48.4}&\textcolor{gray}{89.4}&\textcolor{gray}{77.3}&\textcolor{gray}{92.3}&\textcolor{gray}{42.3}&\textcolor{gray}{13.3}&\textcolor{gray}{77.6}&\textcolor{gray}{29.2}&\textcolor{gray}{59.7}&\textcolor{gray}{86.0}&\textcolor{gray}{54.5}&\textcolor{gray}{51.1}&
    \textcolor{gray}{47.0}&\textcolor{gray}{67.8}&\textcolor{gray}{62.1}&\textcolor{gray}{81.3}&\textcolor{gray}{66.0}\\
    &\textcolor{gray}{spe}&\textcolor{gray}{85.8}&\textcolor{gray}{78.7}&\textcolor{gray}{90.7}&\textcolor{gray}{77.3}&\textcolor{gray}{89.4}&\textcolor{gray}{45.1}&\textcolor{gray}{92.4}&\textcolor{gray}{97.5}&\textcolor{gray}{65.1}&\textcolor{gray}{94.2}&\textcolor{gray}{80.1}&\textcolor{gray}{67.4}&\textcolor{gray}{96.0}&\textcolor{gray}{85.7}&
    \textcolor{gray}{87.8}&\textcolor{gray}{72.6}&\textcolor{gray}{78.9}&\textcolor{gray}{66.0}&\textcolor{gray}{81.3}\\
    &\textcolor{gray}{pre}&\textcolor{gray}{78.1}&\textcolor{gray}{68.3}&\textcolor{gray}{61.6}&\textcolor{gray}{94.4}&\textcolor{gray}{63.0}&\textcolor{gray}{86.5}&\textcolor{gray}{45.8}&\textcolor{gray}{30.8}&\textcolor{gray}{74.2}&\textcolor{gray}{41.2}&\textcolor{gray}{57.8}&\textcolor{gray}{83.1}&\textcolor{gray}{63.2}&\textcolor{gray}{52.7}&
    \textcolor{gray}{56.6}&\textcolor{gray}{51.3}&\textcolor{gray}{70.5}&\textcolor{gray}{86.7}&\textcolor{gray}{56.5}\\
    &\textcolor{gray}{auc}&\textcolor{gray}{89.9}&\textcolor{gray}{84.2}&\textcolor{gray}{79.9}&\textcolor{gray}{89.2}&\textcolor{gray}{89.2}&\textcolor{gray}{86.2}&\textcolor{gray}{85.5}&\textcolor{gray}{76.1}&\textcolor{gray}{79.0}&\textcolor{gray}{74.9}&\textcolor{gray}{79.0}&\textcolor{gray}{86.1}&\textcolor{gray}{87.0}&\textcolor{gray}{78.9}&
    \textcolor{gray}{74.1}&\textcolor{gray}{76.5}&\textcolor{gray}{79.9}&\textcolor{gray}{82.9}&\textcolor{gray}{82.9}\\
    \midrule
    \multirow{2}{*}{\makecell[l]{\textcolor{gray}{FusionM4Net*}\\\textcolor{gray}{\cite{tang2022Fusion}}}}
    &\textcolor{gray}{sen}&\textcolor{gray}{76.3}&\textcolor{gray}{79.5}&\textcolor{gray}{49.5}&\textcolor{gray}{94.4}&\textcolor{gray}{64.0}&\textcolor{gray}{98.1}&\textcolor{gray}{28.8}&\textcolor{gray}{0.3}&\textcolor{gray}{89.7}&\textcolor{gray}{18.8}&\textcolor{gray}{54.8}&\textcolor{gray}{91.4}&\textcolor{gray}{47.7}&\textcolor{gray}{55.3}&
    \textcolor{gray}{40.0}&\textcolor{gray}{76.3}&\textcolor{gray}{76.3}&\textcolor{gray}{95.2}&\textcolor{gray}{44.3}\\
    &\textcolor{gray}{spe}&\textcolor{gray}{83.3}&\textcolor{gray}{81.5}&\textcolor{gray}{90.7}&\textcolor{gray}{64.0}&\textcolor{gray}{94.4}&\textcolor{gray}{29.3}&\textcolor{gray}{96.2}&\textcolor{gray}{99.7}&\textcolor{gray}{50.0}&\textcolor{gray}{98.8}&\textcolor{gray}{89.7}&\textcolor{gray}{66.7}&\textcolor{gray}{97.2}&\textcolor{gray}{89.7}&
    \textcolor{gray}{92.5}&\textcolor{gray}{78.0}&\textcolor{gray}{76.6}&\textcolor{gray}{44.3}&\textcolor{gray}{95.2}\\
    &\textcolor{gray}{pre}&\textcolor{gray}{74.8}&\textcolor{gray}{71.6}&\textcolor{gray}{62.2}&\textcolor{gray}{91.8}&\textcolor{gray}{72.7}&\textcolor{gray}{84.1}&\textcolor{gray}{53.6}&\textcolor{gray}{50.0}&\textcolor{gray}{69.9}&\textcolor{gray}{69.2}&\textcolor{gray}{70.8}&\textcolor{gray}{83.6}&\textcolor{gray}{67.7}&\textcolor{gray}{62.7}&
    \textcolor{gray}{64.5}&\textcolor{gray}{58.5}&\textcolor{gray}{72.6}&\textcolor{gray}{82.3}&\textcolor{gray}{77.0}\\
    &\textcolor{gray}{auc}&\textcolor{gray}{88.1}&\textcolor{gray}{87.3}&\textcolor{gray}{85.7}&\textcolor{gray}{92.5}&\textcolor{gray}{92.5}&\textcolor{gray}{89.5}&\textcolor{gray}{87.9}&\textcolor{gray}{81.3}&\textcolor{gray}{83.1}&\textcolor{gray}{82.5}&\textcolor{gray}{84.9}&\textcolor{gray}{87.2}&\textcolor{gray}{89.6}&\textcolor{gray}{84.6}&
    \textcolor{gray}{78.2}&\textcolor{gray}{82.5}&\textcolor{gray}{84.5}&\textcolor{gray}{83.0}&\textcolor{gray}{83.0}\\
    \bottomrule
  \end{tabular}
\end{table}
\end{landscape}

\begin{figure*}[!htbp]
\centering
\begin{subfigure}{0.32\textwidth}
    \includegraphics[width=\textwidth]{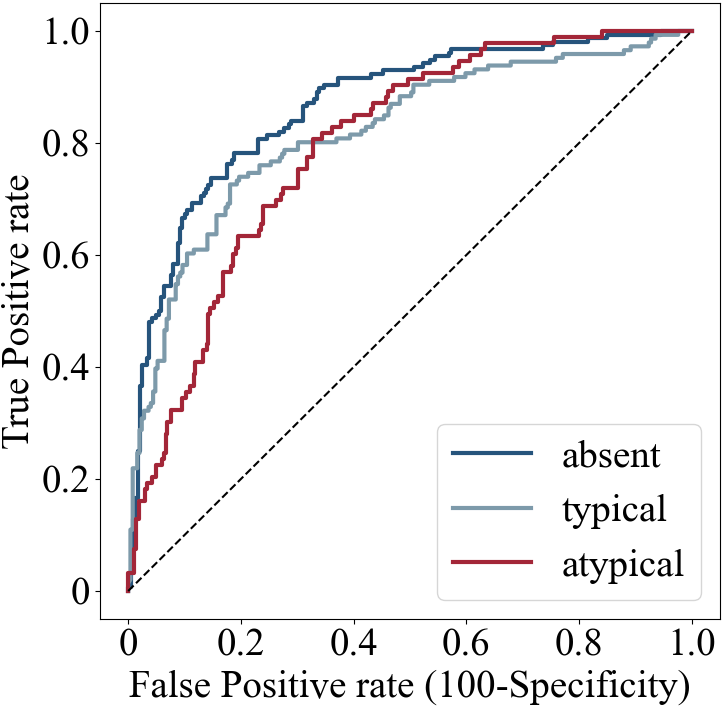}
    \caption{ROC curve of attribute pn.}
\end{subfigure}
\hfill
\begin{subfigure}{0.32\textwidth}
    \includegraphics[width=\textwidth]{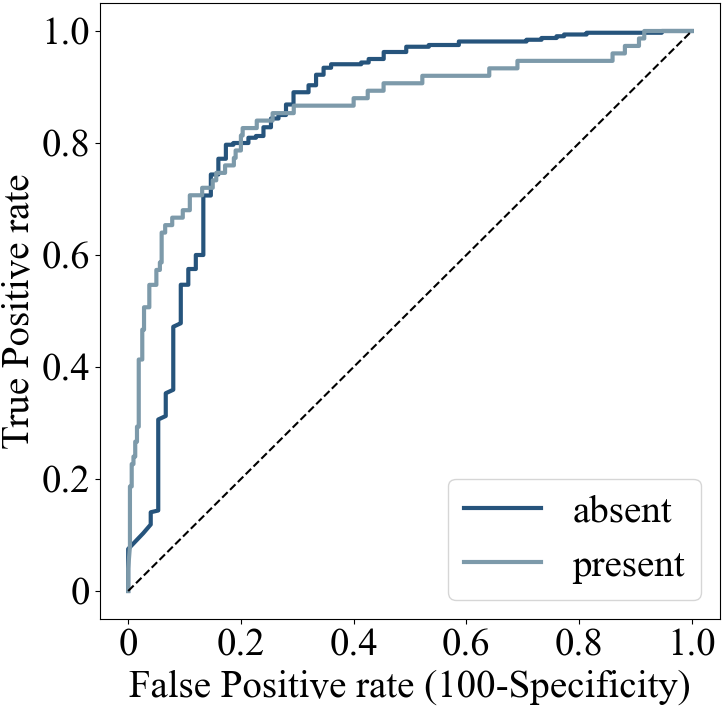}
    \caption{ROC curve of attribute bmv.}
\end{subfigure}
\hfill
\begin{subfigure}{0.32\textwidth}
    \includegraphics[width=\textwidth]{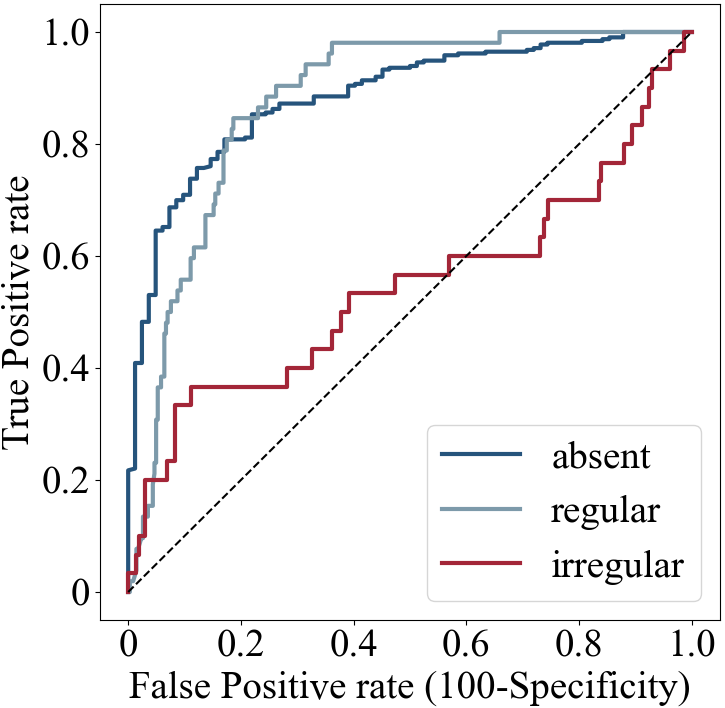}
    \caption{ROC curve of attribute vs.}
\end{subfigure}
\hfill
\begin{subfigure}{0.32\textwidth}
    \includegraphics[width=\textwidth]{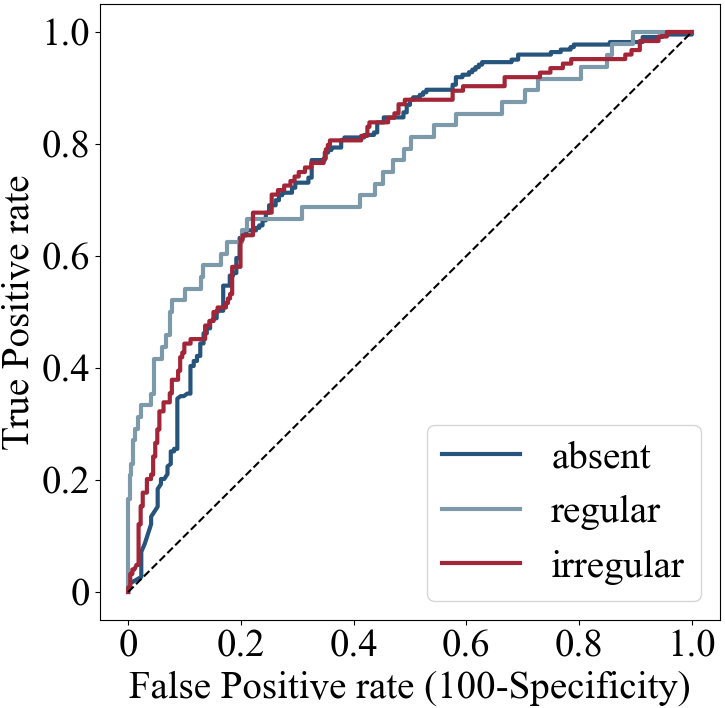}
    \caption{ROC curve of attribute pig.}
\end{subfigure}
\hfill
\begin{subfigure}{0.32\textwidth}
    \includegraphics[width=\textwidth]{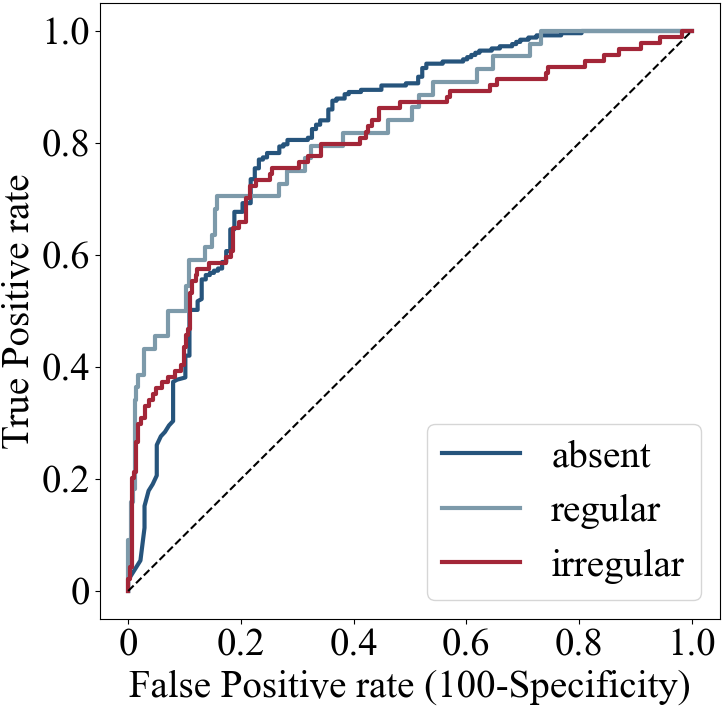}
    \caption{ROC curve of attribute str.}
\end{subfigure}
\hfill
\begin{subfigure}{0.32\textwidth}
    \includegraphics[width=\textwidth]{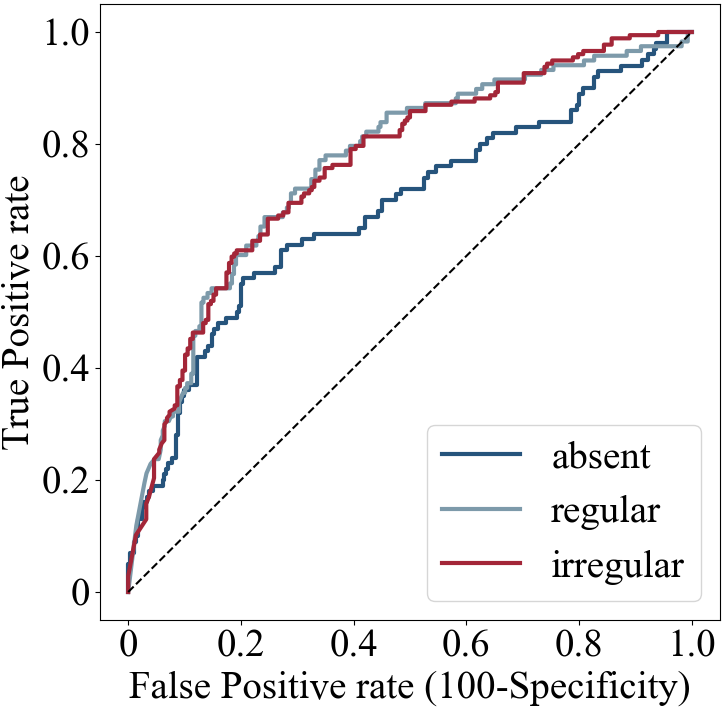}
    \caption{ROC curve of attribute dag.}
\end{subfigure}
\hfill
\begin{subfigure}{0.32\textwidth}
    \includegraphics[width=\textwidth]{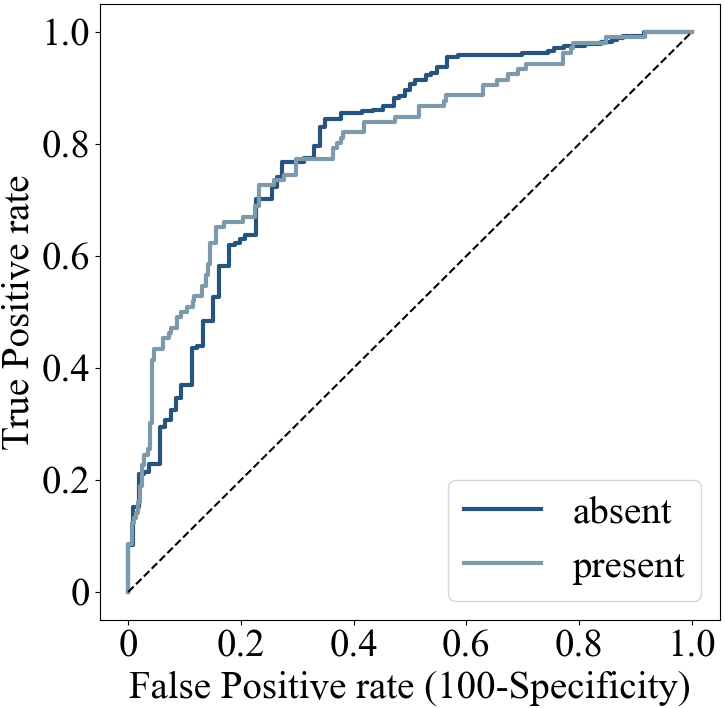}
    \caption{ROC curve of attribute rs.}
\end{subfigure}
\hfill
\begin{subfigure}{0.32\textwidth}
    \includegraphics[width=\textwidth]{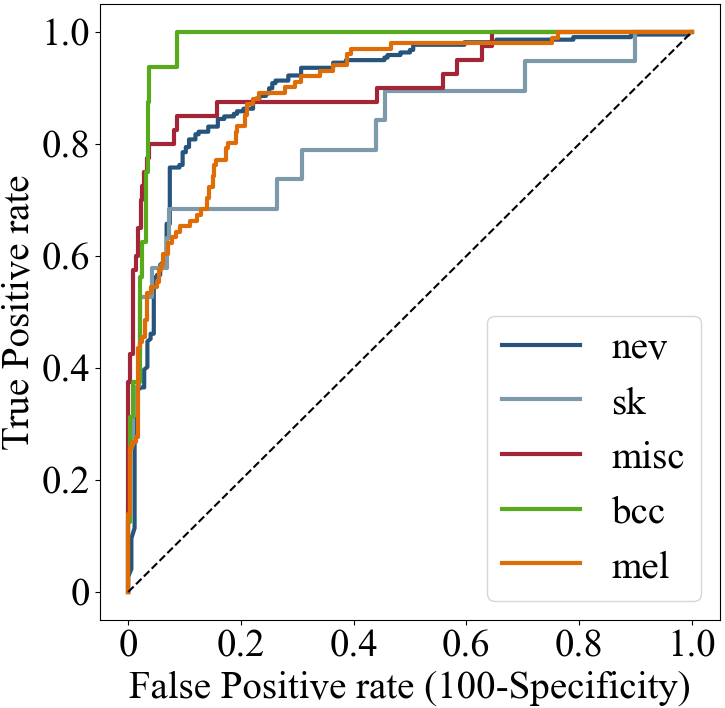}
    \caption{ROC curve of attribute diag.}
\end{subfigure}
\caption{\textbf{derm7pt performance} Receiver Operating Characteristic Curve OvR.}
\label{fig:rocs}
\end{figure*}

\end{document}